%% file: main.tex
\documentclass{article}

\usepackage{arxiv}

\usepackage{amsfonts}
\usepackage{amsmath,bm}
\usepackage{amssymb}
\usepackage{lineno}
\usepackage{tabularx}
\usepackage{multirow}
\usepackage{multicol}
\usepackage{subfig}
\usepackage{graphicx}
\usepackage{caption}
\usepackage{tikz}
\usepackage{color}
\usepackage{xspace}
\usepackage{enumerate}
\usepackage{xurl}
\usepackage[colorlinks,allcolors=blue]{hyperref}
\usepackage{xspace}
\usepackage[normalem]{ulem}
\usepackage{algorithm}%
\usepackage{algorithmicx}%
\usepackage{algpseudocode}%
\usepackage{booktabs}
\usepackage{transparent}

\usepackage[acronym, shortcuts]{glossaries}

\newcommand{\mat}[1]{\mbox{\fontencoding{T1}\sffamily\slshape{#1\/}}} 
\renewcommand{\vec}[1]{\mbox{\textbf{#1}} }

\newcommand{\func}[1]{{\mbox{\usefont{OT1}{pzc}{m}{it}{#1}}}}
\newcommand{\set}[1]{\mathcal{#1}}

\renewcommand{\mathbf}[1]{%
    \pdfliteral direct {2 Tr 0.3 w} 
     #1%
    \pdfliteral direct {0 Tr 0 w}%
}

\newcolumntype{L}{>{\raggedright\arraybackslash}X}
\newcolumntype{C}{>{\centering\arraybackslash}X}
\newcolumntype{R}{>{\raggedleft\arraybackslash}X}
\newcommand{\std}[1]{{\scalebox{0.4}{\transparent{0.7}{$\pm #1$}}}}
\newcommand{\best}[1]{\pdfliteral direct {2 Tr 0.3 w}#1\pdfliteral direct {0 Tr 0 w}} 
\newcommand{\textbest}[1]{\pdfliteral direct {2 Tr 0.3 w}#1\pdfliteral direct {0 Tr 0 w}} 
\newcommand{\secondbest}[1]{{\underline{{#1}}}}
\newcommand{\textsecondbest}[1]{{\underline{{#1}}}}

\newacronym{eo}{EO}{Earth Observation}
\newacronym{mvl}{MVL}{Multi-View Learning}
\newacronym{sits}{SITS}{Satellite Image Time Series}
\newacronym{f1}{F1}{F1 Macro}
\newacronym{prs}{PRS}{Performance Robustness Score}
\newacronym{ds}{DS}{Deformation Score}
\newacronym{cr}{CR}{Class change Ratio}
\newacronym{r2}{$R^2$}{Coefficient of Determination}
\newacronym{lfmc}{LFMC}{Live Fuel Moisture Content}
\newacronym{publicyield}{CYP}{Crop Yield Prediction}
\newacronym{cropharvestB}{CropH-b}{CropHarvest binary}
\newacronym{cropharvestM}{CropH-m}{CropHarvest multi}
\newacronym{pm25}{PM25}{Particulate Matter 2.5}
\newacronym{mlp}{MLP}{Multi-Layer Perceptron}
\newacronym{tempcnn}{TempCNN}{Temporal CNN}
\newacronym{maug}{MAug}{Missing data as Augmentation}
\newacronym{tempd}{TempD}{Temporal Dropout}
\newacronym{sensd}{SensD}{Sensor Dropout}
\newacronym{allmissing}{CoM}{Combinations of Missing views}

\newcommand{\bad}[1]{#1}

\title{Missing Data as Augmentation in the Earth Observation Domain: A Multi-View Learning Approach}

\date{} 					

\author{ 
    Francisco Mena$^{1,2}$\href{https://orcid.org/0000-0002-5004-6571}{\includegraphics[scale=0.06]{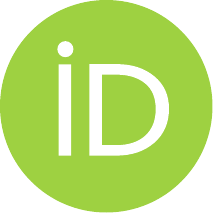}\hspace{1mm}},
	Diego Arenas$^{2}$\href{https://orcid.org/0000-0001-7829-6102}{\includegraphics[scale=0.06]{imgs/orcid.pdf}\hspace{1mm}}, 
	Andreas Dengel$^{1,2}$\href{https://orcid.org/0000-0002-6100-8255}{\includegraphics[scale=0.06]{imgs/orcid.pdf}\hspace{1mm}} \\
	$^1$Department of Computer Science,	University of Kaiserslautern-Landau (RPTU), Germany;\\
	$^2$SDS, German Research Center for Artificial Intelligence (DFKI), Germany.\\
	\texttt{f.menat@rptu.de}, \{\texttt{diego.arenas, andreas.dengel}\}\texttt{@dfki.de} \\
}

\renewcommand{\headeright}{A Preprint}
\renewcommand{\shorttitle}{Missing as Augmentation in Earth Observation}

\newcommand*\citep[1]{\cite{#1}}
\newcommand*\citet[1]{\cite{#1}}

\begin{document}
\maketitle

\begin{abstract}
Multi-view learning (MVL) leverages multiple sources or views of data to enhance machine learning model performance and robustness. 
This approach has been successfully used in the Earth Observation (EO) domain, where views have a heterogeneous nature and can be affected by missing data.
Despite the negative effect that missing data has on model predictions, the ML literature has used it as an augmentation technique to improve model generalization, like masking the input data.
Inspired by this, we introduce novel methods for EO applications tailored to MVL with missing views.
Our methods integrate the combination of a set to simulate all combinations of missing views as different training samples.
Instead of replacing missing data with a numerical value, we use dynamic merge functions, like average, and more complex ones like Transformer. 
This allows the MVL model to entirely ignore the missing views, enhancing its predictive robustness.
We experiment on four EO datasets with temporal and static views, including state-of-the-art methods from the EO domain.
The results indicate that our methods improve model robustness under conditions of moderate missingness, and improve the predictive performance when all views are present. 
The proposed methods offer a single adaptive solution to operate effectively with any combination of available views.
\end{abstract}


\input{content/introduction}
\input{content/relatedwork}
\input{content/methods}
\input{content/experiments}
\input{content/discussion}
\input{content/conclusion}

\section*{Acknowledgments}
F. Mena acknowledge support through a scholarship of the University of Kaiserslautern-Landau.

\section*{Data and code availability}
The data used in this manuscript corresponds to public benchmark datasets released in \citet{tseng2021crop,rao2020sar,pm25}. We provide functions to facilitate the processing of these to a machine learning ready structure. 
This is available on our GitHub at \url{https://github.com/fmenat/com-views}.

\bibliographystyle{apalike} 
\bibliography{main}

\clearpage
\appendix
\input{content/supp}

\end{document}

%% file: content/introduction.tex
\section{Introduction} \label{sec:introduction}

Nowadays, the usage of multiple data sources, sensors, or views in ML models has become a standard practice in various applications and domains \citep{yan2021deep}. The reason is that by using multiple sources of information, the individual data can be ratified and complemented to enhance predictive models \citep{hong2021more,saintefaregarnot2022multi}. 
\gls{eo} is one of the domains where \gls{mvl} has been used to provide comprehensive insights in various applications \citep{mena2024common}.
In our work, we refer to a \textit{view} as all features in a specific data source.
For example, a view can be an optical or radar \gls{sits}, weather conditions, topographic information, or various metadata, consisting of a heterogeneous scenario with different spatio-temporal resolutions.
Thus, there are \textit{temporal views}, with multi-temporal data and \textit{static views}, with single-date data.
This diversity distinguishes research done in \gls{eo} from other ML domains such as vision and text \cite{rolf2024mission}.
Furthermore, \gls{eo} views might not be a persistent source of information as researchers commonly assume.

The \gls{eo} domain faces challenges due to the finite lifespan of remote sensors, noise, and cloudy conditions in optical sensors \cite{shen2015missing}. 
Besides, unexpected errors can affect the availability of the data, such as the failure of the Sentinel-1B satellite in 2021.
This problem leads to scenarios with missing data, which hinder accurate predictions and introduce biases in ML models \cite{choi2019embracenet,hong2021more,saintefaregarnot2022multi}.
For instance, \citet{mena2024igarss} evidence the negative predictive impact that missing views have in  different vegetation applications, highlighting the lack of robustness of different \gls{mvl} models. 
Even current ML models, like Transformers, are not naturally robust to missing views \cite{ma2022multimodal,tseng2023lightweight,chen2024novel}.
This leaves open questions such as how to increase the robustness of \gls{mvl} models to missing views.

Despite the negative impact of missing data in ML models, many studies actively incorporate it during learning.
For instance, it has been used as a masking operator for self-supervision in text \cite{devlin2018bert}, signal \cite{cao2018brits} and vision domains \cite{he2022masked}.
Another option is to use \gls{maug} techniques, such as the dropout layer \cite{srivastava2014dropout}, and augmentation operations in the vision domain (e.g. crop).
In the \gls{eo} domain, the \gls{maug} technique has been used in \gls{mvl} models.
One case is to drop all features from a random view (sensor) during training \cite{wang2022self,mena2024notyet}.
Another case corresponds to including all combinations of missing views, as experimented by \citet{gawlikowski2023handling} in out-of-distribution detection. 
However, most of these works use a fixed-size merge function (like concatenation) and validate on \gls{eo} datasets with only static views. 

In this work, we introduce methods based on two main components, all \gls{allmissing} and a dynamic merge function at feature-level. 
The \gls{allmissing} acts as an augmentation technique, simulating all combinations of missing views in the training set, which relates to literature \cite{gawlikowski2023handling,mena2024notyet}. 
However, we apply the \gls{maug} technique at the feature-level instead of at the input \cite{mena2024notyet} and use dynamic merge instead of fixed functions \cite{gawlikowski2023handling}.
We find that integrating the \gls{allmissing} with a merge function that ignores the missing views enhances the predictions and robustness of \gls{mvl} models. 
The dynamic merge can be a simple average or a more complex function. Inspired by literature on using ML models as aggregators \cite{lee2019set,bugueno2020learning,wang2021multiview,guo2024skysense}, we include some alternative functions based on gated fusion, cross-attention fusion, and memory fusion.

We validate the proposed methods on four \gls{eo} datasets with temporal and static views, and compare them to five state-of-the-art methods in the \gls{eo} domain. For assessing the robustness to missing data, we simulate missing views during inference and compare the predictions to full-view data.
The evidence suggests that our methods have better robustness than competing methods in cases of moderate missingness, and in some cases improve predictive performance when all views are available. 

Overall, our main contributions are as follows: i) we propose a \gls{maug} technique tailored to missing views in \gls{mvl} at feature level (\gls{allmissing}), ii) adapt the \gls{maug} technique to ignore the features of the missing views by using a dynamic fusion, and iii) validate on \gls{eo} datasets with classification and regression tasks, considering temporal and static views.
The proposed methods, inspired by sensor invariant modeling \cite{francis2024sensor,mena2024notyet}, allow a model to generate adaptive predictions based on the available views. 
Our code will be released at \url{https://github.com/fmenat/com-views}.

%% file: content/relatedwork.tex
\section{Related work} \label{sec:related}

\paragraph{MVL with EO data.}
Recently, there has been an increase in \gls{eo} research using multiple data sources to enhance ML model predictions \cite{camps2021deep}. The main difference in the \gls{mvl} models investigated is how the data is fused \cite{mena2024common}. 
Input-level fusion has been the common choice for this, i.e. merge the data before feeding a ML model. 
For instance, \citet{kussul2017deep} feed a CNN model with just the concatenation of different sensors (multi-spectral and radar images) for land-use classification, while \citet{ghamisi2016hyperspectral} concatenate specialized hand-crafted features from hyper-spectral and LiDAR images.
However, several works have shown that learning view-dedicated feature extractors (encoders) improves results \cite{hong2021more,saintefaregarnot2022multi,ferrari2023fusing,mena2024adaptive}.
For example, \citet{audebert2018rgb} propose a \gls{mvl} model for multi-spectral and topographic images that fuses across multiple layers of CNN encoders. 
Later, \citet{zhang2020hybrid} show that including a fusion in the decision layer (as a hybrid fusion) improves the results for land-use segmentation.
Additionally, \citet{ofori-ampofo2021crop} evidence that using specialized encoder architectures for optical and radar \gls{sits} benefits the feature-level fusion in a crop-type classification use-case.
Furthermore, there have been efforts in exploring geospatial foundational models using multi-view data. 
For instance, Presto \cite{tseng2023lightweight} uses an input-level fusion with a Transformer model, while SkySense \cite{guo2024skysense} uses a feature-level fusion based on view-dedicated models that are fine-tuned based on the available views in the downstream tasks.

\paragraph{Missing data in EO applications}
Different forms of missing (such as errors and anomalies) are present in \gls{eo} data \cite{shen2015missing}.
As expected, when the missing information in the input data increases (at spectral, spatial, or temporal dimensions), the predictions of ML models get worse \cite{fasnacht2020robust,ofori-ampofo2021crop,ebel2023uncrtaints}.
In addition, specific data sources are more relevant than others. 
For example, the lack of an optical view critically affects models' accuracy \cite{hong2021more,mena2024igarss}. 
Nevertheless, there is evidence that when a view is missing, training with additional views can supplement and increase the model robustness \cite{inglada2016improved,hong2021more}. 
For instance, \citet{ferrari2023fusing} and \citet{saintefaregarnot2022multi} show that when optical images are missing in \gls{sits} due to cloudy conditions, placing the fusion far away from the input layer increases the model robustness. 
Furthermore, in \citet{mena2024igarss} three techniques that mitigate the effect of missing views in \gls{mvl} are compared. 
The first is to impute the missing view with a numerical value. The second replaces the missing view with a similar sample in the training set. The last one ignores the missing views in the aggregation through a dynamic fusion.
The latter is the technique that has shown greater robustness in various \gls{eo} datasets when views are missing \cite{mena2024igarss}, as well when images are missing in \gls{sits} \cite{che2024linearly}. 
Furthermore, modifying model components can also increase predictive robustness, as has been shown when sharing layer weights \cite{zheng2021deep,mena2024notyet}.

\paragraph{Leverage missing data} 
Simulate missing data has been progressively used in ML research, from standard augmentation techniques (\gls{maug}) in the vision domain (like crops) to masking out image patches for reconstruction, as a self-supervised framework \cite{he2022masked}. 
In the natural language domain, it has been used to learn token embeddings from reconstruction tasks \cite{devlin2018bert}.
In the signal domain, it has been studied how to best impute data in time series \cite{cao2018brits, du2023saits}.
In the \gls{eo} domain, masking out input data and learning to reconstruct it has been widely used for self-supervised learning, such as in SatMAE \cite{cong2022satmae}, SITS-Former \cite{yuan2022sits}, Presto \cite{tseng2023lightweight}, and OmniSat \cite{astruc2024omnisat}. 
Similarly, the dropout operator has not only considered data augmentation \cite{bouthillier2015dropout} but used as such.
For instance, \citet{fasnacht2020robust} introduce a spectral dropout to increase the model robustness to missing spectral bands for hyper-spectral image segmentation.
In the same predictive task, \citet{haut2019hyperspectral} use spatial dropout (random occlusion) as a \gls{maug} technique.
Furthermore, randomly dropping images in \gls{sits}, i.e. \gls{tempd}, has been presented as an effective \gls{maug} technique for prediction \cite{saintefaregarnot2022multi}. 
Recently, the \gls{maug} has been applied to sensors, i.e. \gls{sensd}, as a way to learn inter-sensor representations \cite{wang2022self}, to avoid overfitting to dominant sensors \cite{chen2024novel}, to assess sensors contribution \cite{ekim2024deep}, and to increase model robustness to missing data \cite{mena2024notyet}.
Additionally, \citet{gawlikowski2023handling} show that the \gls{maug} applied to views can be used for out-of-distribution detection in two-view image classification tasks.

Most of the works that use the concept of \gls{maug} in EO focus on masking out the data at the input-level or using a fixed merge function. 
In practice, this means imputing the missing features with a \textit{fake} value to obtain a fixed-size input in all scenarios. 
Moreover, the literature positions the input-level as a fusion strategy with low predictive capacity and robustness.
To overcome these disadvantages, our work focuses on using a \gls{maug} at feature-level by ignoring the missing views in the \gls{mvl} model.
We achieve this by using dynamic merge functions, as we show below.

%% file: content/methods.tex
\section{Multi-view learning with missing data} \label{sec:methods}

\begin{figure*}[t]
\centering
\includegraphics[width=\textwidth, page=1]{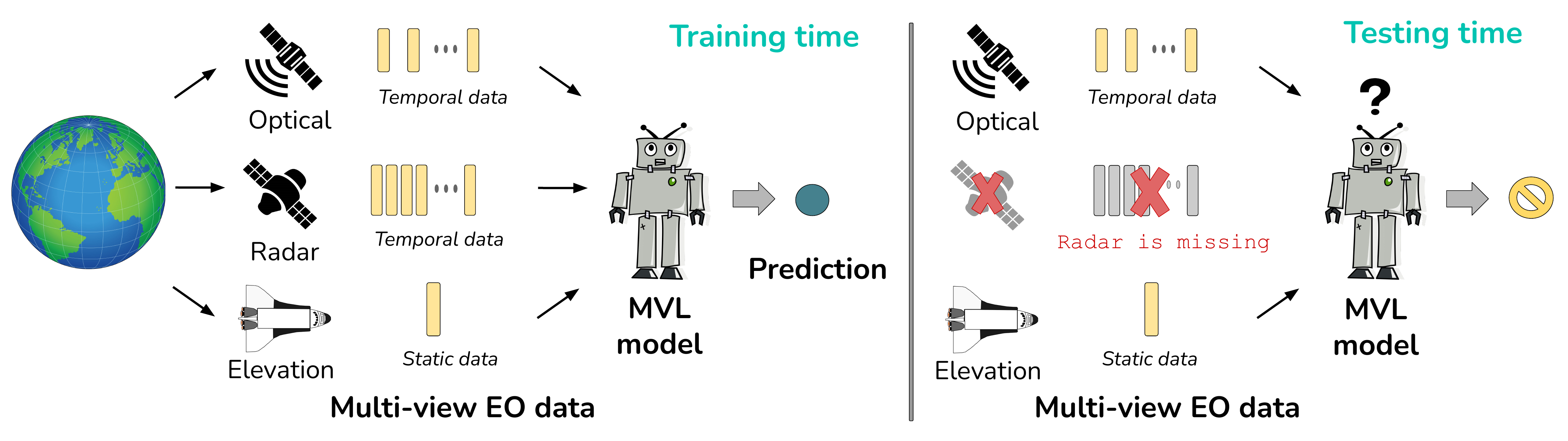}
\caption{Illustration of a \gls{mvl} scenario with three views available during training, while at inference time, one view is missing.} \label{fig:problem}
\end{figure*}

\subsection{Problem notation}
Given the multi-view input data for a sample $i$, $\set{X}^{(i)} = \{ \mat{X}_v^{(i)} \}_{v \in \set{V}}$, with $\set{V}$ the set of all views, the objective is to find a \gls{mvl} model $\func{G}(\cdot)$ that approximates the corresponding target $y^{(i)}$, i.e. $\hat{y}^{(i)} = \func{G}(\set{X}^{(i)})$. 
The views $\mat{X}_v$ can be temporal (time-series data) or static (single-date data).
The learning is through minimizing a loss function of the form $\func{L}(y^{(i)}, \func{G}(\set{X}^{(i)}))$, over a training set of $N$ samples, $\set{D} = \{ \set{X}^{(i)}, y^{(i)} \}_{i=1}^N$.
In the case of missing views, instead, we observe $\tilde{\set{X}}^{(i)} = \{ \mat{X}_v^{(i)} \}_{v \in \set{V}^{(i)}}$, with $\set{V}^{(i)} \subseteq \set{V}$, the set of available views for a sample $i$.
Then, the number of views is $m = |\set{V}|$ and $m^{(i)} = |\set{V}^{(i)}|$, with $m^{(i)}> 0$. 
As views in \gls{mvl} models could be any set of features, we consider a view as all features from a data source (e.g. optical, radar, weather).
We consider a full-view training scenario, with expected missing views at inference, as illustrated in Figure~\ref{fig:problem}. 

\subsection{Basis of multi-view learning}

\paragraph{Input-level fusion}
This fusion strategy directly merges the input features of the views.  As \gls{eo} views have different (spatio-temporal) resolutions, an alignment step is required to match all the dimensions: 
$\mat{X}_F^{(i)} = \func{concat} ( \func{alignment}( \set{X}^{(i)} ) )$.
Then, these merged features are fed to a single ML model: $\hat{y}^{(i)} = \func{G}( \mat{X}_F^{(i)})$.
However, in this fusion strategy, there is no clear way to deal with missing views, $\tilde{\set{X}}^{(i)}$. 
For instance, \citet{hong2021more} present a zero-imputation of the missing data, i.e. $\set{X}^{(i)} = \tilde{\set{X}}^{(i)} \cup \left\{\mat{0}\right\}_{v \in \set{V} \setminus \set{V}^{(i)}}$. 
Subsequent research on \gls{maug} has used the zero-imputation in the missing features \cite{gawlikowski2023handling,mena2024notyet}, i.e. augment the training data by masking out views with a zero value.  
Nonetheless, zero is an arbitrary value that creates bias depending on data normalization and transformations applied. 

\paragraph{Feature-level fusion}
To avoid forcing a view-alignment, and have a single model that handles the multi-view information, this strategy extracts high-level features through view-dedicated encoders: $\vec{z}_v^{(i)} = \func{G}_v^{\text{enc}} ( \mat{X}_v^{(i)} ) \ \forall v \in \set{V}$. 
In addition, a normalization layer (with learnable parameters) is used in each encoder to scale and harmonize the different representations.
Then, a merge function combines this information, obtaining a joint representation, $\vec{z}_F^{(i)} = \func{M} ( \{  \vec{z}_v^{(i)} \}_{v \in \set{V}} )$. The merge function $\func{M}(\cdot)$ can take any form, such as concatenation or dynamic functions.
Then, a prediction head is used to obtain the final prediction: $\hat{y}^{(i)} = \func{G}^{head} ( \vec{z}_F^{(i)} )$.
In the following, we explain how to handle missing views in the latter fusion strategy.

\subsection{Dynamic feature-level fusion} \label{sec:methods:dynamic}
Inspired by permutation \cite{lee2019set} and sensor \cite{francis2024sensor,mena2024notyet} invariant models, we rely on ignoring the encoded features associated to the missing views. For this, the \gls{mvl} model encodes and merges only the available views:
\begin{align}
   \label{eq:encoders_forward} \vec{z}_v^{(i)} &= \func{G}_v^{\text{enc}} ( \mat{X}_v^{(i)} ) \ \ \forall v \in \set{V}^{(i)}  , \\
   \label{eq:dynamic_fusion} \vec{z}_F^{(i)} &= \func{M} \left( \left\{  \vec{z}_v^{(i)} \right\}_{v \in \set{V}^{(i)}} \right),
\end{align}
with $\vec{z}_v^{(i)} \in \mathbb{R}^d$ and $\vec{z}_F^{(i)} \in \mathbb{R}^{d^F}$.
However, when the fused dimension ($d^F$) depends on the number of views ($m^{(i)}$), as with concatenation, fusion cannot be dynamic. 
Thus, we use merge functions that yield the same fused dimension regardless of the fused views, i.e $d^F = d$. We call these dynamic merge functions.
A simple case is a linear combination with the same weight, i.e. average as
\begin{equation}
    \func{M} \left( \mat{Z}^{(i)} \right) = \frac{1}{m^{(i)}} \sum_{v \in \set{V}^{(i)}} \vec{z}_v^{(i)} \ .
\end{equation} 
Furthermore, we present some alternatives in the following.

\paragraph{Gated fusion}
Instead of using the same weight for all views as in the average, we use a data-driven weighted fusion \cite{choi2019embracenet,mena2024adaptive}. Considering the encoded features from all views as $\mat{Z}^{(i)} = \func{stack} ( \{  \vec{z}_v^{(i)} \}_{v \in \set{V}} ) \in \mathbb{R}^{m \times d}$,
the gated merge function is expressed by 
\begin{equation}\label{eq:gated_fusion}
    \func{M} \left( \mat{Z}^{(i)} \right)  = \sum_{v \in \set{V}} \func{softmax}\left( \mat{A}^{(i)} \right)_v^{\top} \odot  \mat{Z}_v^{(i)} \ ,
\end{equation}
with $\mat{A}^{(i)}$ the fusion weights.
Then, instead of modeling a single fusion weight
for all dimensions $d$ in each view, $\mat{A}^{(i)} \in \mathbb{R}^{1\times m}$ \cite{mena2024adaptive}, we use a per-dimension weight in each view, i.e. $\mat{A}^{(i)} \in \mathbb{R}^{d \times m}$, calculated as 
\begin{equation}\label{eq:gated_calc}
\mat{A}^{(i)} = \mat{W}_{\text{G}} \cdot \func{flatten} \left( \mat{Z}^{(i)} \right) + \vec{b} \ ,
\end{equation}
with $\mat{W}_{\text{G}}$ and $\vec{b}$ learnable parameters.
In the case of missing data, the fusion weights of the unavailable views are modified, such as $\func{softmax}(\vec{A}^{(i)})_{v} = \vec{0}  \ \forall v \in \set{V} \setminus \set{V}^{(i)}$. With this adaptation of weights, the features of the missing views are ignored in the merge Eq.~\eqref{eq:gated_fusion}. 
As the fusion weights require all views to be calculated, Eq.~\eqref{eq:gated_calc}, and we only forward over the available views, Eq.~\eqref{eq:encoders_forward}, during implementation we impute the missing features with zeros, $\vec{z}_v^{(i)}  = \vec{0} \ \forall v \in \set{V} \setminus \set{V}^{(i)}$.

\paragraph{Cross-attention fusion}  
Inspired by Transformer layers used to fuse \gls{eo} data \cite{ma2022multimodal,chen2024novel}, we use a learnable parameter, called \textit{fusion token}, $\vec{f} \in \mathbb{R}^d$ to query the multi-view data. Consider the encoded features from the available views with the fusion token as $\mat{Z}^{(i)} = \func{stack} ( \vec{f}, \{  \vec{z}_v^{(i)} \}_{v \in \set{V}^{(i)}} ) \in \mathbb{R}^{(1+m^{(i)}) \times d}$, the cross-attention merge function is expressed by 
\begin{equation}\label{eq:cross_fusion}
\func{M} \left( \mat{Z}^{(i)} \right)  = \func{softmax} \left( \mat{A}^{(i)} \right)_0 \cdot \mat{Z}^{(i)} \mat{W}_{\text{V}}    \ ,
\end{equation}
with $\mat{A}^{(i)} \in \mathbb{R}^{(1+m^{(i)}) \times (1+m^{(i)})}$ the cross-view (and token) attention weights, and  $\mat{W}_{\text{V}}$ a learnable parameter. The values $\mat{A}_0^{(i)} \in \mathbb{R}^{1 \times (1+m^{(i)})}$ are the view-attention weights of the fusion token used to aggregate the views. These weights are computed by a self-attention mechanism as follows
\begin{equation}\label{eq:cross_calc}
\mat{A}^{(i)} =  \mat{Z}^{(i)} \mat{W}_{\text{Q}} \cdot  \mat{Z}^{(i)}\mat{W}_{\text{K}} + \mat{b}  \ ,
\end{equation}
with $\mat{W}_{\text{Q}}$, $\mat{W}_{\text{K}}$, and $\mat{b}$ learnable parameters
As the matrix computation in Eq.~\eqref{eq:cross_calc} depends exclusively on the available views, the model naturally avoids attending the missing views when merging Eq.~\eqref{eq:cross_fusion}, i.e.  $\mat{A}_{v}^{(i)} = \vec{0}  \ \ \forall v \in \set{V} \setminus \set{V}^{(i)}$.
We use a multi-head mechanism and stacked layers to increase the learning of cross-view features \cite{vaswani2017attention}.
In contrast to previous works \cite{lee2019set,chen2024novel}, we include a view-specific positional encoding.

\paragraph{Memory fusion} 
Inspired by RNN models used to fuse multi-view EO data \cite{wang2021multiview}, we employ a memory-based fusion. 
The memory is updated one view at a time with an empty initial memory, expressed by 
\begin{equation}\label{eq:memory_calc}
\vec{h}_v^{(i)} = \func{R} \left( \vec{z}^{(i)}_v, \vec{h}_{v-1}^{(i)} \right)  \ \ \ \vec{h}_{0} = \vec{0},
\end{equation}
with $\func{R}$ a RNN model, and $v \in \{1, \ldots, m^{(i)} \}$.
Then, the memory-based fused vector corresponds to
$
\func{M} (\{  \vec{z}_v^{(i)} \}_{v \in \set{V}^{(i)}} )  = \vec{h}_{m^{(i)}}^{(i)}$.
This means that the fused vector is the memory (or hidden state in the RNN model) after being recursively updated with all views. 
As the recursive operation Eq.~\eqref{eq:memory_calc} and fused vector is invariant to the number of views given as input, it naturally ignores missing views.
Similar to the cross-attention fusion, we stack multiple LSTM layers in $\func{R}(\cdot)$ to increase the learning of cross-view features.
Since RNNs are order-dependent, a random permutation can be used to avoid a bias in the order in which the views are given. 
However, using the proposed \gls{maug} technique is enough for generalization, as shown in the appendix.

\subsection{All combinations of missing views} \label{sec:methods:maug}

As previous works have shown, randomly dropping views during training increases the model robustness to missing views \cite{tseng2023lightweight,chen2024novel}. 
However, it can negatively affect the model accuracy in the full-view scenario \cite{mena2024notyet}.
Thus, we consider augmenting the training samples by modeling all combinations of missing views at feature-level.
Then, assuming a full-view training set, the augmented features extracted from the $i$-th sample are $\{ \{ \vec{z}_v^{(i)} \}_{v \in \set{V}^{(j)}} \}_{\set{V}^{(j)} \in \mathcal{T}} $, with $\mathcal{T}= \{ \set{V}^{(j)}:  \set{V}^{(j)} \subseteq \set{V},  \set{V}^{(j)} \neq \emptyset \}$ the augmented list. 
Here, the number of possible combinations is the same as the power set of $\set{V}$ minus the no-view case, i.e. $|\mathcal{T}| = 2^{m} -1 $. 
For instance, for optical, radar, and weather views, the augmented list is $\mathcal{T}=$\{(optical/radar/weather), (optical/radar), (optical/weather), (radar/weather),  (optical), (radar), (weather)\}.
We named this \gls{maug} technique as \textbf{\acrfull{allmissing}}.
The usage of \gls{allmissing} during training is illustrated in Algorithm~\ref{pseudo:permu}.

\begin{algorithm}[!t]
\caption{\gls{allmissing} technique at feature-level}\label{pseudo:permu}
\begin{algorithmic}[1]
\Require $\mathcal{D}: \{ \set{X}^{(i)}, y^{(i)} \}_{i=1}^N$ - multi-view dataset 
\Require $\func{G}(\cdot)$ - initialized \gls{mvl} model
\Require $\mathcal{T}$ - set of all missing views cases
\Ensure $\func{G}(\cdot)$ - Trained \gls{mvl} model 
\For{$( \set{X}^{(i)}, y^{(i)} ) \in \mathcal{D}$} 
    \State Obtain $\vec{z}_v^{(i)}$ by forwarding over all view-encoders as Eq.~\eqref{eq:encoders_forward}
    \State Initialize $\mathcal{Y}^{(i)}$ as an empty list
    \For{$\set{V}^{(j)} \in \mathcal{T}$}
        \State Obtain $\vec{z}_{F_j}^{(i)}$ from the available views $\set{V}^{(j)}$ with the dynamic function as Eq.~\eqref{eq:dynamic_fusion} 
        \State Obtain $\hat{y}^{(i)}_j$ by applying $\func{G}^{head} ( \vec{z}_{F_j}^{(i)} ) $
        \State Update $\mathcal{Y}^{(i)}$ by attaching the prediction $\hat{y}^{(i)}_j$
    \EndFor
    \State Calculate the loss function based on $y^{(i)}$ and $\hat{y}^{(i)}_j \in \mathcal{Y}^{(i)}$ as Eq.~\eqref{eq:loss}
    \State Update $\func{G}(\cdot)$ by gradient descent
\EndFor
\end{algorithmic}
\end{algorithm}
We consider a balanced contribution between the full-view and missing views predictive performance. This means that all augmented samples from $\mathcal{T}$ have the same weight in the loss function during training, expressed by
\begin{equation} \label{eq:loss}
   \ell = \frac{1}{N} \sum_{i=1}^N  \frac{1}{|\mathcal{T}|} \sum_{\set{V}^{(j)} \in \mathcal{T}} \func{L}(y^{(i)}, \hat{y}^{(i)}_j ) \ ,
\end{equation}
with $ \hat{y}^{(i)}_j= \func{G}\left(\{ \mat{X}_v^{(i)} \}_{v \in \set{V}^{(i)}}\right)$.
Furthermore, to reduce computational operations of the multiple predictions, we forward over the encoders (biggest computation bottleneck) only once, while the fusion and prediction are done $|\mathcal{T}|$ times, see Algorithm~\ref{pseudo:permu} for details.

%% file: content/experiments.tex
\section{Experiments} \label{sec:experiments}

\subsection{Datasets}
We use the following pixel-wise \gls{eo} datasets with static and temporal views. More details on the feature description can be found in \ref{sec_app:data}.

\paragraph{\Gls{cropharvestB}} 
We use the CropHarvest dataset for crop recognition with four views \cite{tseng2021crop}. 
This involves a binary task in which the presence of any crop growing at a given location is predicted.
It has 69,800 samples around the globe between 2016 and 2021. 
Each sample has three temporal views: multi-spectral optical (11 bands), radar (2 polarization bands), and weather (2 bands) \gls{sits}.
These time series have one value per month over 1 year.
Besides, the samples have one static view, the topographic information (2 bands).
All features have a pixel resolution of 10 m.
Furthermore, we use a multi-class version, \textbf{\Gls{cropharvestM}}. 
This is a subset of 29,642 samples with 10 crop types to predict (mutually exclusive), and the same input views from \gls{cropharvestB}. 

\paragraph{\Gls{lfmc}}
We use a dataset for moisture content estimation with six views \cite{rao2020sar}. This considers a regression task in which the vegetation water (moisture) per dry biomass (in percentage) in a given location is predicted.
There are 2,578 samples in the western US between 2015 and 2019.
Each sample has two temporal views: multi-spectral optical (8 bands) and radar (3 bands) \gls{sits}. 
These time series have one value per month over 4 months. 
In addition, the samples have four static views: topographic information (2 bands), soil properties (3 bands), canopy height (ordinal feature), and land-cover (12 classes).
All features were interpolated to a pixel resolution of 250 m.

\paragraph{\Gls{pm25}}
We use a dataset for PM2.5 estimation with three views \cite{pm25}.  This involves a regression task in which the concentration of PM2.5 in the air (in $ug/m^3$) in a particular city is predicted.
The dataset has 167,309 samples in five Chinese cities between 2010 and 2015.
Each sample has three temporal views: atmospheric conditions (3 bands), atmospheric dynamics (4 bands), and precipitation (2 bands).
The data are at hourly resolution, of which we keep a 3-day window for the estimation, i.e. signals of 72 time-steps are used as input.

\subsection{Setup and competing methods}

We named our methods using the \gls{allmissing} technique at feature-level as \textbf{F\gls{allmissing}-av} (for average), \textbf{F\gls{allmissing}-ga} (for gated), \textbf{F\gls{allmissing}-cr} (for cross-attention), and \textbf{F\gls{allmissing}-me} (for memory fusion).
For the cross-attention fusion (F\gls{allmissing}-cr), the function consists of one layer with eight heads and 40\% of dropout, while for the memory fusion (F\gls{allmissing}-me), it consists of two bidirectional layers with LSTM units and 40\% of dropout. Variations in the selection of these hyper-parameters are shown in the \ref{sec_app:arch_sel}.

For comparison, we consider the following supervised methods related to the \gls{eo} domain.
Three methods that perform zero-imputation in the missing views. 
Two of these use \gls{maug} techniques at input-level: 
\textbf{I\gls{tempd}-co} \cite{saintefaregarnot2022multi}, using the \gls{tempd} technique, 
and \textbf{I\gls{sensd}-co} \cite{mena2024notyet}, using the \gls{sensd}.
One method that uses \gls{allmissing} at feature-level: \textbf{F\gls{allmissing}l-co} \cite{gawlikowski2023handling}, a model that merges by concatenation with a weighted loss and prediction that we extend to multiple views. 
In addition, we include three methods that ignore the missing views: 
\textbf{F\gls{sensd}-cr}, adapted from images \cite{chen2024novel} to pixel-wise time series, using cross-attention fusion at feature-level with the \gls{sensd} technique, 
\textbf{FEmbr-sa} \cite{choi2019embracenet}, a feature-level fusion method that randomly samples features from different views in the fused representation,
and \textbf{ESensI-av} \cite{mena2024notyet}, a view-invariant model using ensemble aggregation (averaging) without \gls{maug} techniques. 
Since self-supervised methods use a bunch of data outside the training set and more input views, we do not consider them because it would not be a fair comparison.

\paragraph{Implementation} 
We apply a z-score normalization to the input data. 
The categorical and ordinal views (like land-cover and canopy height) are one-hot-vector encoded. 
We use the best encoder architectures for the selected datasets \cite{najjar2024data,mena2024igarss}. This corresponds to a 1D CNN encoder for temporal views, and an MLP encoder for static views. 
We use two layers with 128 units on all encoder architectures, with 20\% of dropout and a final layer normalization.
After fusion, we use a MLP with one layer as the prediction head.
For optimization, we use ADAM optimizer with a batch-size of 128 and early stopping with a patience of five. The stopping criterion is applied over the full-view prediction.
The loss function is cross-entropy in classification and mean squared error in regression tasks. 
We use a weight in the loss function (inverse to the number of samples in each class) to balance the class unbalance in classification.
For competing methods, we use 30\% of dropout in I\gls{tempd}-co and the no ratio version in the I\gls{sensd}-co method \cite{mena2024notyet}.

\paragraph{Evaluation} 
We use 10-fold cross-validation repeated three times to reduce results variability.
We simulate missing views in the validation fold, as illustrated in Figure~\ref{fig:problem}.
We experiment with different degrees of \textit{missingness}: i) moderate missingness, when only one view is missing, ii) extreme missingness, when all views are missing except one. 
We include the results with no missing views (full-view scenario) for reference.
For assessing the predictive performance, we use the \gls{f1} in classification and the \gls{r2} in regression tasks.

\subsection{Results with missing views}

In order to assess the effect when a single view is missing, or it is available, we decide to select a few views that individually are the most effective for the task. For this, we train a model individually on each view to predict the task and select the two views that reach the best predictions. We refer to these are the \textit{top} views. 
For \gls{cropharvestB}, \gls{cropharvestM}, and \gls{lfmc} these are optical and radar views, as observed in the literature \cite{hong2021more,zheng2021deep,mena2024igarss}, while for \gls{pm25} are dynamic and condition views. The results from all the auxiliary views are in the Table~\ref{tab:individual} in the \ref{sec_app:ind}.

\begin{table*}[!b] 
\caption{Predictive performance in the classification tasks (\acrshort{f1} score) for different cases of missing views (moderate and extreme). The \textbest{best} and \textsecondbest{second best} value are highlighted. In parentheses is the number of available views. }\label{tab:missing:f1}
\scriptsize
\begin{tabularx}{\textwidth}{l|cCCCC|cCCCC} \cmidrule{2-11}
\multicolumn{1}{c}{} & \multicolumn{5}{c}{\textbf{\acrfull{cropharvestB}}} & \multicolumn{5}{c}{\textbf{\acrfull{cropharvestM}}} \\ 
\toprule
& (4/4) No & \multicolumn{2}{c}{(3/4) Only missing} & \multicolumn{2}{c|}{(1/4) Only available} & (4/4) No & \multicolumn{2}{c}{(3/4) Only missing} & \multicolumn{2}{c}{(1/4) Only available} \\
Method & Missing & \multicolumn{1}{c}{Radar}  & Optical & \multicolumn{1}{c}{Optical} & Radar & Missing & \multicolumn{1}{c}{Radar}  & Optical & \multicolumn{1}{c}{Optical} & Radar
\\ 
\midrule
I\gls{tempd}-co           & $0.817 \std{ 0.008} $ & $0.798 \std{ 0.009} $ & $0.717 \std{ 0.013} $    & $0.668 \std{ 0.015} $        & $0.506 \std{ 0.075} $ & $0.635 \std{0.017} $ & $0.552 \std{0.019} $ & $0.336 \std{0.021} $       & $0.364 \std{0.029} $            & $0.154 \std{0.017} $ \\
I\gls{sensd}-co           & 	$0.787 \std{0.015} $ &	$0.780 \std{0.015} $ &	$0.747 \std{0.022} $ &	$0.765 \std{0.02} $ &	$0.572 \std{0.092} $ & $0.608 \std{0.013} $ &	$0.587 \std{0.016} $ &	$0.437 \std{0.038} $ &	$0.587 \std{0.017} $ &	$0.207 \std{0.052} $  \\
F\gls{sensd}-cr          & $0.776 \std{0.045} $ &	$0.748 \std{0.091} $ &	$0.707 \std{0.106} $ &	$0.569 \std{0.21} $ &	$0.527 \std{0.169} $  & $0.559 \std{0.126} $ &	$0.451 \std{0.232} $ &	$0.302 \std{0.173} $ &	$0.367 \std{0.259} $ &	$0.196 \std{0.169} $ \\
F\gls{allmissing}l-co           & $0.832 \std{0.007} $	& $\secondbest{0.827} \std{0.006} $	 & $\secondbest{0.804} \std{0.006} $ & 	$0.787 \std{0.010} $	& ${0.686} \std{0.010}$ & $0.650 \std{0.020} $ &	$0.614 \std{0.019} $ &	$0.543 \std{0.014} $ &	$0.601 \std{0.019} $ &	$0.388 \std{0.015} $  \\
FEmbr-sa & $0.821 \std{0.007} $ &	$0.817 \std{0.007} $ &	$0.786 \std{0.007} $ &	$0.764 \std{0.008} $ &	$0.676 \std{0.012} $ &  $0.633 \std{0.015} $ &	$0.598 \std{0.022} $ &	$0.478 \std{0.015} $ &	$0.543 \std{0.024} $ &	$0.312 \std{0.012} $ \\
ESensI-av           & $0.802 \std{ 0.011} $ & $0.806 \std{0.010} $  & $0.767 \std{ 0.012} $ & ${0.791} \std{ 0.013} $        & $\best{0.701} \std{ 0.012} $ & $0.605 \std{0.019} $ & $0.577 \std{0.021} $ & $0.478 \std{0.015} $   &  $\secondbest{0.635} \std{0.023} $            & $\best{0.444} \std{0.022} $ \\ 
\midrule
F\gls{allmissing}-av     & $\secondbest{0.837} \std{ 0.005} $ & $\best{0.834} \std{ 0.005} $ & $\secondbest{0.804} \std{0.007} $ & $\best{0.809} \std{ 0.006} $        & $0.615 \std{ 0.040} $ & $\best{0.686} \std{0.018} $ & $\best{0.648} \std{0.020} $  & $\secondbest{0.549} \std{0.011} $     & $\best{0.649} \std{0.018} $            & $0.428 \std{0.017} $  \\
F\gls{allmissing}-ga   & $\best{0.839} \std{0.005} $ &	$\best{0.834} \std{0.006} $ &	$\best{0.810} \std{0.008} $ &	$\secondbest{0.805} \std{0.006} $ &	$0.681 \std{0.018} $ & $\secondbest{0.678} \std{0.016} $ &	$\secondbest{0.642} \std{0.015} $ &	$\best{0.566} \std{0.015} $ &	$0.595 \std{0.139} $ &	$0.399 \std{0.093} $ \\
F\gls{allmissing}-cr    & $0.826 \std{0.007} $ &	$0.823 \std{0.007} $ &	$0.799 \std{0.007} $ &	$0.801 \std{0.008} $ &	$\secondbest{0.693} \std{0.008} $ & $0.660 \std{0.019} $ &	$0.637 \std{0.019} $ &	$\secondbest{0.549} \std{0.013} $ &	$0.632 \std{0.017} $ &	$\secondbest{0.439} \std{0.015} $  \\ 
F\gls{allmissing}-me    & $0.836 \std{0.006} $ &	$0.818 \std{0.007} $ &	$0.800 \std{0.007} $ &	$0.800 \std{0.008} $ &	$0.681 \std{0.014} $  & $0.670 \std{0.015} $ &	$0.633 \std{0.016} $ &	$0.535 \std{0.011} $ &	$0.629 \std{0.015} $ &	$0.422 \std{0.013} $  \\
\bottomrule
\end{tabularx}
\end{table*}

\begin{table*}[!t]    
\caption{Predictive performance in the regression tasks (\acrshort{r2} score) for different cases of missing views (moderate and extreme). The \textbest{best} and \textsecondbest{second best} values are highlighted. In parentheses is the number of available views. The symbol $\dagger$ represent values below -10.}\label{tab:missing:r2}
\scriptsize
\begin{tabularx}{\textwidth}{l|cCCCC|cCCCC} \cmidrule{2-11}
\multicolumn{1}{c}{} & \multicolumn{5}{c}{\textbf{\acrfull{lfmc}}} & \multicolumn{5}{c}{\textbf{\acrfull{pm25}}} \\
\toprule
& (6/6) No & \multicolumn{2}{c}{(5/6) Only missing} & \multicolumn{2}{c|}{(1/6) Only available} & (3/3) No & \multicolumn{2}{c}{(2/3) Only missing} & \multicolumn{2}{c}{(1/3) Only available} \\
Method & Missing & \multicolumn{1}{c}{Radar}  & Optical & \multicolumn{1}{c}{Optical} & Radar  & Missing & \multicolumn{1}{c}{Condition}  & Dynamic & \multicolumn{1}{c}{Dynamic} & Condition \\ 
\midrule
I\gls{tempd}-co           & $\secondbest{0.691} \std{0.052} $  & $0.036 \std{0.100} $    & $0.036 \std{0.100} $     & -$\bad{0.036} \std{0.06} $   & -${0.036} \std{0.06} $ &  $\best{0.866} \std{0.093} $   & $0.074 \std{0.035} $     & -$\bad{0.124} \std{0.14} $                  & $0.073 \std{0.035} $            & -$\bad{0.124} \std{0.14} $  \\
I\gls{sensd}-co           & $0.546 \std{0.053} $ &	$0.537 \std{0.047} $ &	$0.349 \std{0.053} $ &	$0.315 \std{0.057} $ &	$0.121 \std{0.049} $ & $0.511 \std{0.058} $ &	$0.319 \std{0.065} $ &	$0.083 \std{0.056} $ &	$0.318 \std{0.066} $ &	$0.083 \std{0.056} $  \\
F\gls{sensd}-cr          & $0.433 \std{0.178} $ &	$0.383 \std{0.243} $ &	$0.123 \std{0.147} $ &	-$0.031 \std{0.557} $ &	-$0.264 \std{0.47} $ & $0.187 \std{0.289} $ &	-$1.046 \std{2.77} $ &	-$0.971 \std{5.10} $ &	-$1.091 \std{2.78} $ &	-$1.202 \std{5.07} $  \\
F\gls{allmissing}l-co           & -$\bad{0.643} \std{2.14} $ &	-$\bad{0.505} \std{2.17} $ &	-$\bad{0.434} \std{2.00} $ &	-$\bad{0.661} \std{2.13} $ &	-$\bad{1.172} \std{1.93} $  & -$\bad{0.316} \std{0.30} $ &	$0.157 \std{0.186} $ &	-$\bad{0.010} \std{0.12} $ &	$\secondbest{0.358} \std{0.391} $ &	$0.125 \std{0.104} $   \\
FEmbr-sa &  $0.559 \std{0.065} $ &	$0.532 \std{0.060} $ &	$0.270 \std{0.058} $ &	-$5.300 \std{1.80} $ &	-$0.796 \std{0.52} $ & -$1.366 \std{4.39} $ &	-$0.389 \std{2.51} $ &	-$0.135 \std{0.42} $ & $\dagger$ &	-$1.077 \std{1.52} $ \\ 
ESensI-av           & $0.321 \std{0.038} $  & $0.294 \std{0.036} $  & $0.239 \std{0.035} $   &  ${0.194} \std{0.224} $    & -$\bad{0.130} \std{0.36} $ &   $0.231 \std{0.077} $      & $0.255 \std{0.059} $     & $\secondbest{0.069} \std{0.106} $                 & ${0.334} \std{0.078} $            & $0.034 \std{0.135} $    \\
\midrule
F\gls{allmissing}-av    & $0.628 \std{0.051} $  & ${0.606} \std{0.066} $  & ${0.435} \std{0.058} $    & -$\bad{9.633} \std{12.90} $    & -$\bad{7.425} \std{7.24} $ &  $\secondbest{0.660} \std{0.103} $ &	$\best{0.556} \std{0.162} $ &	-$\bad{0.441} \std{0.65} $ &	-$\bad{6.570} \std{3.81} $ &	$\best{0.409} \std{0.147} $  \\
F\gls{allmissing}-ga   & $\best{0.700} \std{0.043} $ &	$\best{0.683} \std{0.044} $ &	$\best{0.491} \std{0.05} $ &	$\secondbest{0.326} \std{0.09} $ &	$\secondbest{0.165} \std{0.094} $ & $0.046 \std{0.151} $ &	$0.096 \std{0.133} $ &	$\best{0.135} \std{0.122} $ &	$0.239 \std{0.345} $ &	$\secondbest{0.237} \std{0.155} $ \\
F\gls{allmissing}-cr    & $0.425 \std{0.154} $ &	$0.139 \std{1.187} $ &	$0.227 \std{0.153} $ &	-$2.403 \std{4.533} $ &	-$2.292 \std{3.112} $ &  $0.414 \std{0.108} $ &	$\secondbest{0.374} \std{0.126} $ &	-$0.130 \std{0.222} $ &	-$0.279 \std{2.271} $ &	-$0.203 \std{0.717} $ \\
F\gls{allmissing}-me   & $0.644 \std{0.066} $ &	$\secondbest{0.641} \std{0.06} $ &	$\secondbest{0.479} \std{0.061} $ &	$\best{0.330} \std{0.072} $ &	$\best{0.169} \std{0.077} $ & -$0.033 \std{0.172} $ &	$0.079 \std{0.133} $ &	-$0.004 \std{0.137} $ &	$\best{0.430} \std{0.087} $ &	$0.230 \std{0.141} $ \\
\bottomrule
\end{tabularx}
\end{table*}
In Table \ref{tab:missing:f1} we compare the \gls{f1} score of the methods in different cases of missing views for the classification tasks.
When there are no missing views or moderate missingness, the best results are obtained by our methods, with F\gls{allmissing}-av and F\gls{allmissing}-ga. However, in extreme cases, our methods become comparable to competing methods.
In this extreme missingness, the ESensI-av method effectively handles missing views, competing with our methods and outperforming when only radar view is available.
On the other hand, we notice that the methods based on \gls{sensd} and \gls{tempd} are highly affected by missing views, i.e. less robust.

For the regression tasks, we display the \gls{r2} in Table \ref{tab:missing:r2}.
Here, the impact of missing views is more severe than in classification, with most methods reaching negative \gls{r2} values in extreme missingness.
This difficulty in regression is expected \cite{mena2024igarss} as methods predict a continuous value which disperses in different magnitudes, while in classification the change is only binary, see Sec.~\ref{sec:results:shift} for a visual illustration.
Nonetheless, some of our methods are robust enough and obtain the best results in extreme cases, e.g. F\gls{allmissing}-me.
Moreover, our methods (F\gls{allmissing}-av and F\gls{allmissing}-ga) effectively handle the moderate missingness in both datasets. 
Although the I\gls{tempd}-co has the best results in the full-view scenario of \gls{pm25} data, it has poor robustness, strongly decreasing performance when views are missing.
This outcome is inverse for F\gls{allmissing}-ga, F\gls{allmissing}-me, and F\gls{allmissing}l-co methods, suggesting that these learned missing views scenarios better than full-view ones.
Furthermore, leaving the fusion to a random view-selection, as it is the case of FEmbr-sa, does not work in the \gls{pm25} data, as it reaches negative \gls{r2} values in all cases.

Overall, we observe that the F\gls{allmissing}-av, despite using the simple aggregation function between the compared ones, performs the best without missing data and in moderate missingness. 
For the extreme missingness, the best combination for merge function in the \gls{allmissing} depends on the task type. In classification, the best predictions are obtained with F\gls{allmissing}-av and F\gls{allmissing}-cr, while in regression is with F\gls{allmissing}-ga and F\gls{allmissing}-me.
Furthermore, some of our methods are greatly affected in these extreme cases. 
For instance, the F\gls{allmissing}-av method shows a great performance decline in regression.
In addition, the baseline F\gls{allmissing}l-co has poor results in most cases of regression, reflecting the poor transferability to other tasks and ineffectiveness of the zero-imputation as observed in the literature \cite{mena2024notyet,che2024linearly}.

\subsection{Results with a fraction of missing views}

 \begin{figure*}[t!]
\centering
\subfloat
{\includegraphics[width=0.475\textwidth]{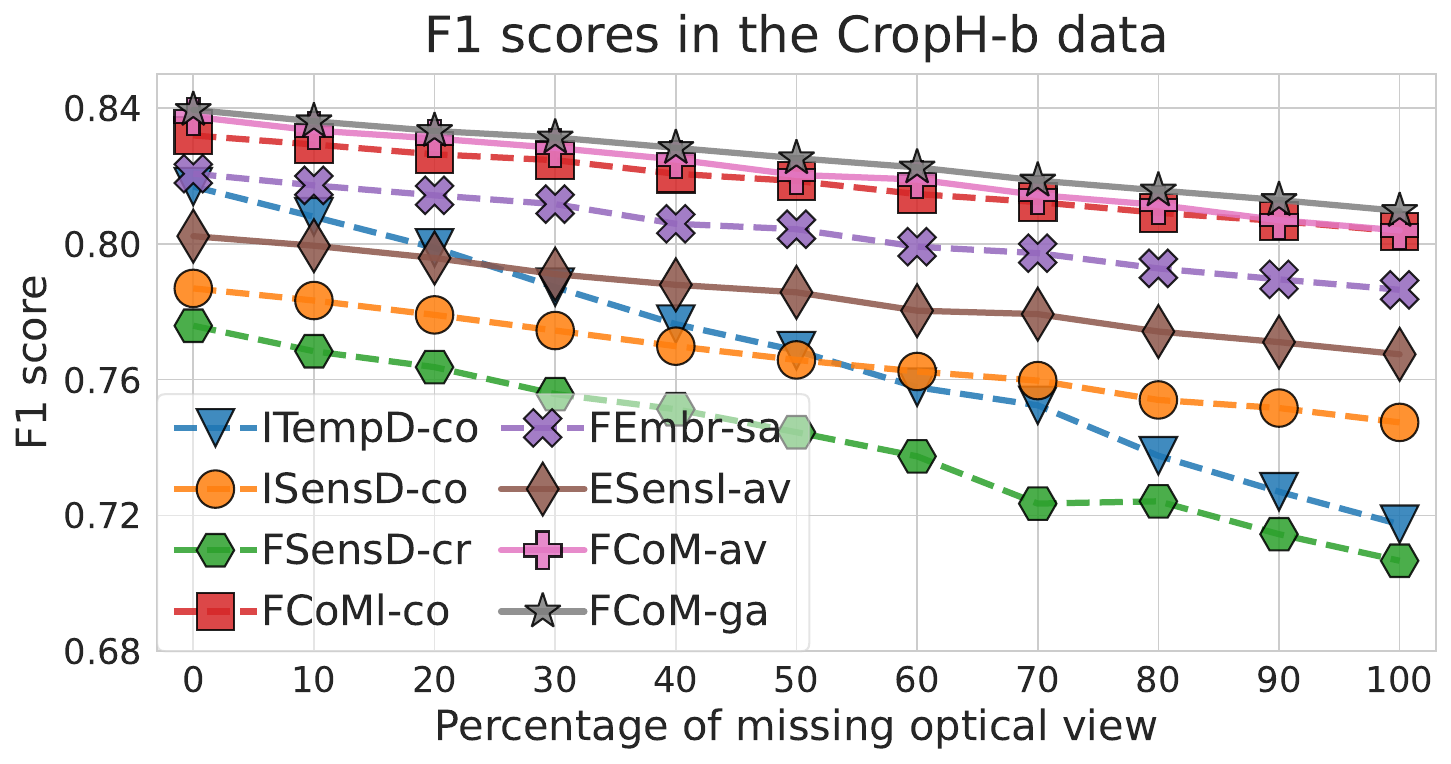}}
\hfill
\subfloat
{\includegraphics[width=0.475\textwidth]{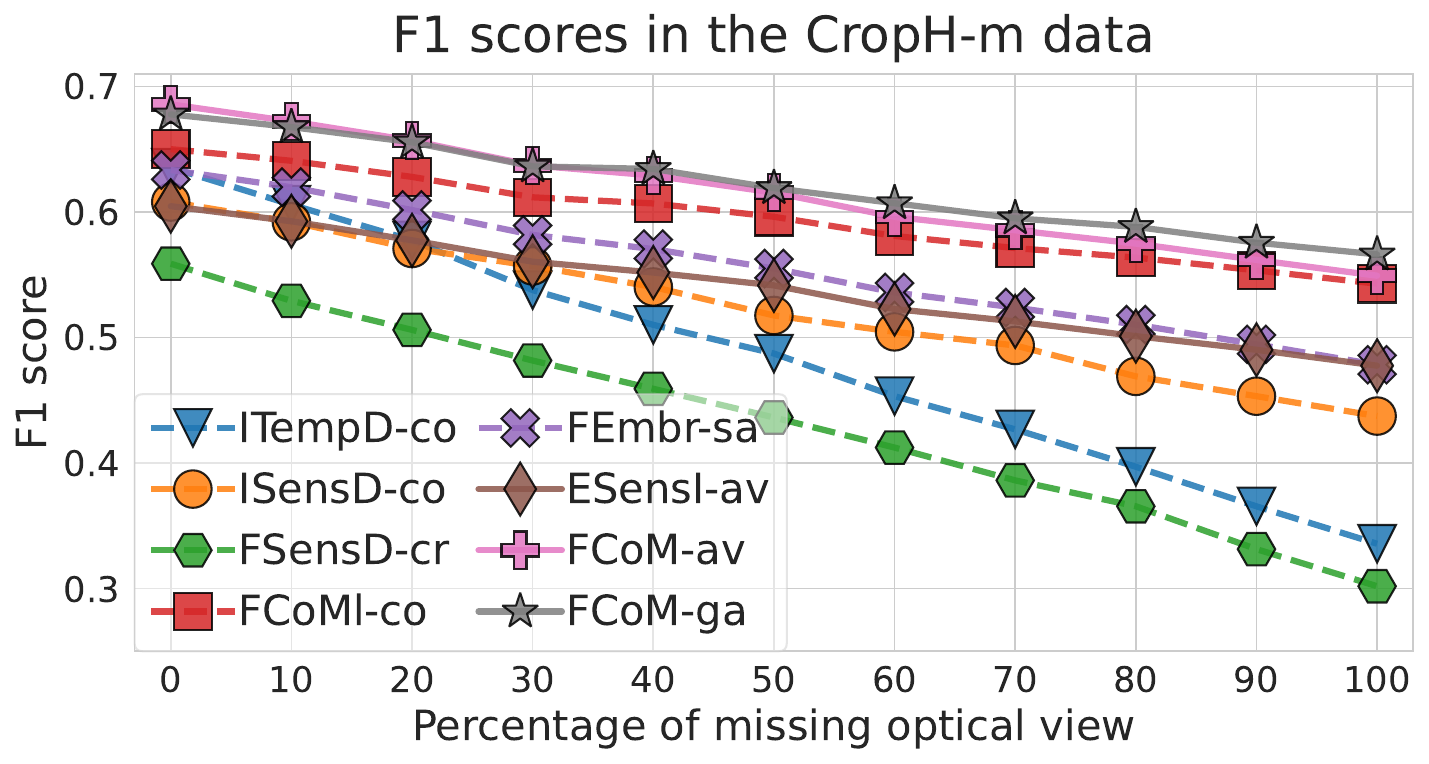}}\\
\subfloat
{\includegraphics[width=0.475\textwidth]{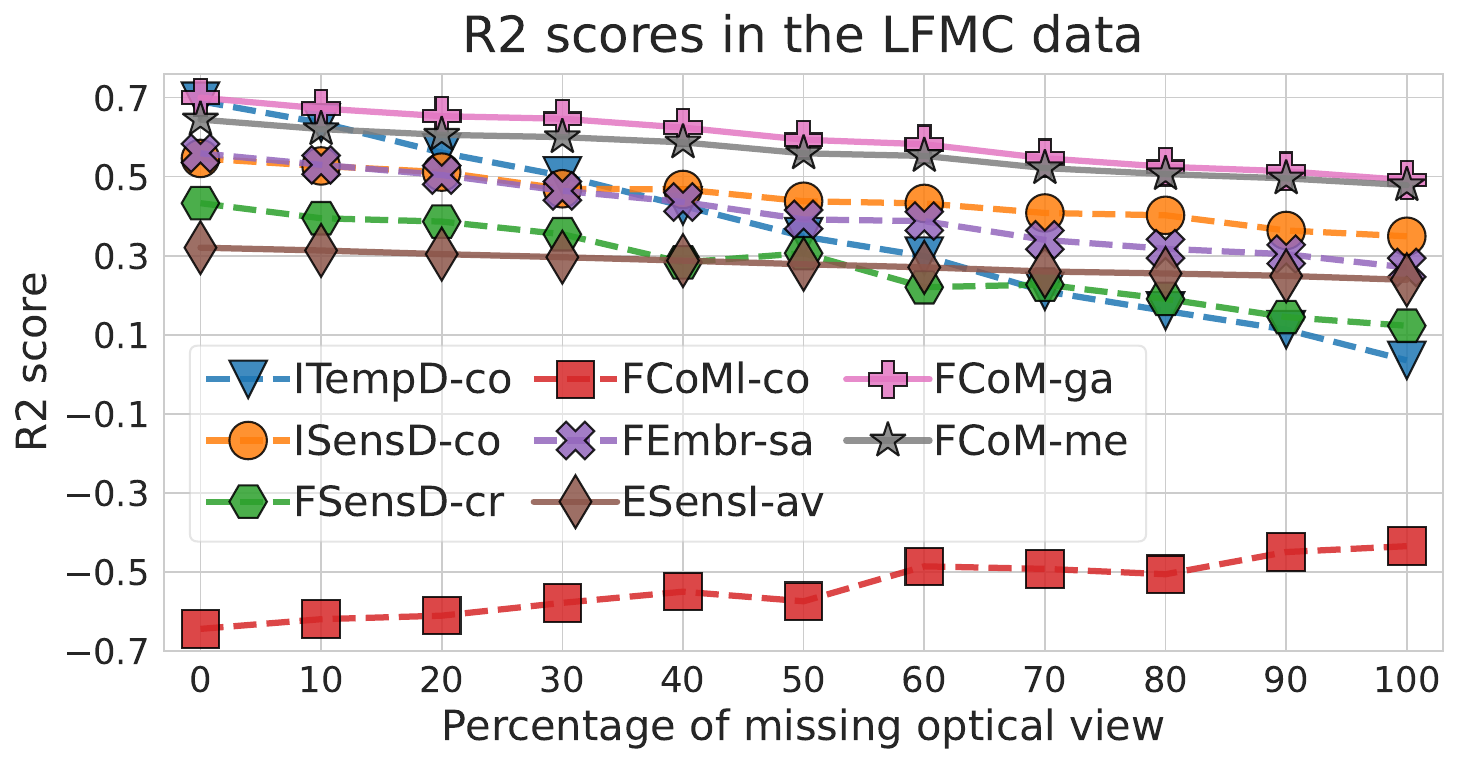}}
\hfill
\subfloat
{\includegraphics[width=0.475\textwidth]{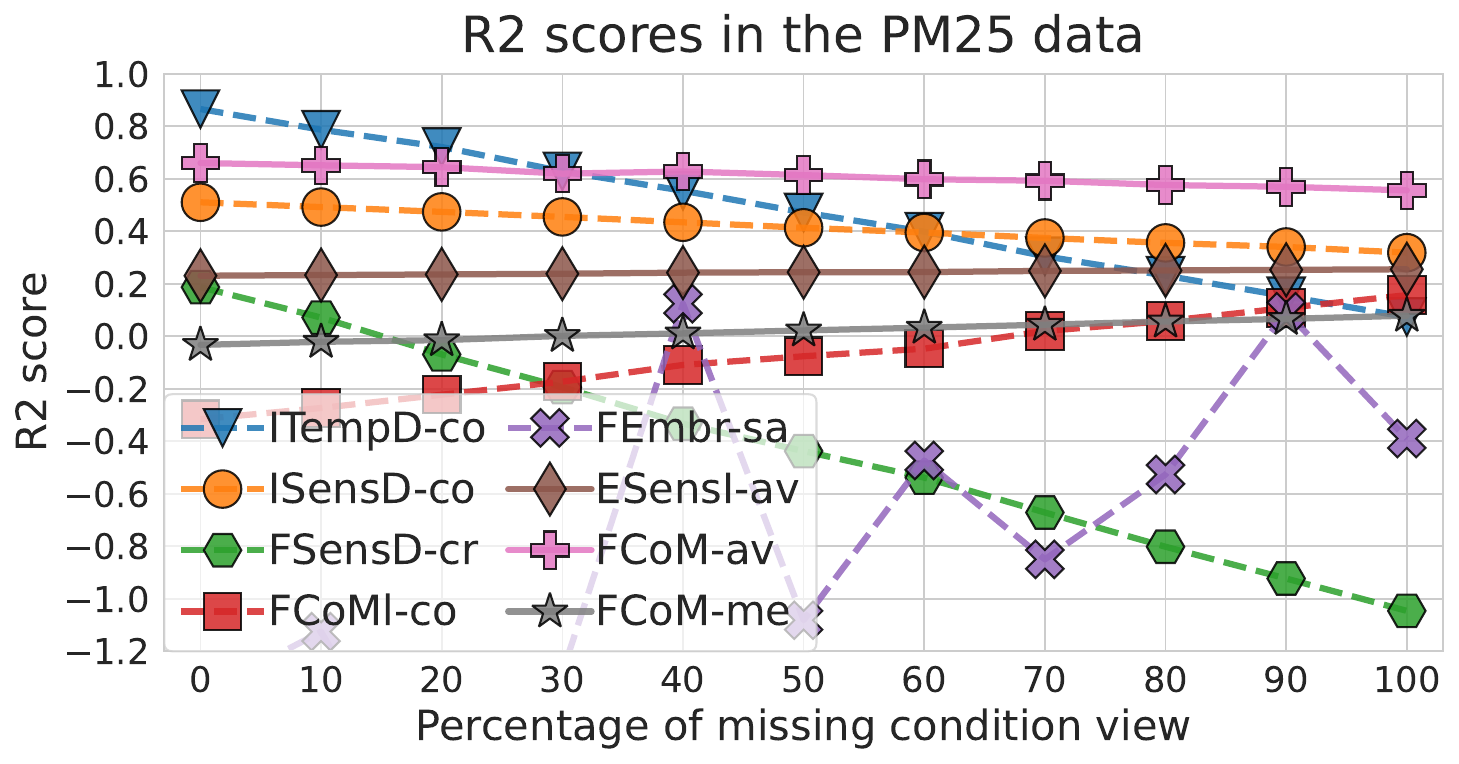}}
\caption{Predictive performance when varying percentages of validation samples have a \textit{top} view missing.}\label{fig:perc_missing}
\end{figure*} 
In Figure \ref{fig:perc_missing} we display the predictive performance when one top view for prediction is missing in some samples. We include the results when the other top view is missing in the \ref{sec_app:another_view}.
We present our two best methods in each dataset.
In classification, our F\gls{allmissing}-ga method has the best results along the percentages of missing views, competing with F\gls{allmissing}-av and F\gls{allmissing}l-co.
However, the competing method F\gls{allmissing}l-co has a curious behavior in regression, focusing only on the missing views cases and obtaining poor results overall. The FEmbr-sa method also has a strange behavior in the \gls{pm25} data, being ineffective for this scenario.
In the \gls{pm25} data, the I\gls{tempd}-co method has the best results until 30\% of the samples have the condition view missing, from there, the F\gls{allmissing}-av becomes the best. This is due to the greater robustness of our method.
Overall, we notice that our methods have the best robustness behavior when the number of samples with missing views increases. This behavior, in the performance curve, corresponds to a good balance between a small slope and a high value.

\subsection{Prediction shift due to missing views} \label{sec:results:shift}

\begin{figure}[t!]
\centering
\subfloat
{\includegraphics[width=0.475\textwidth]{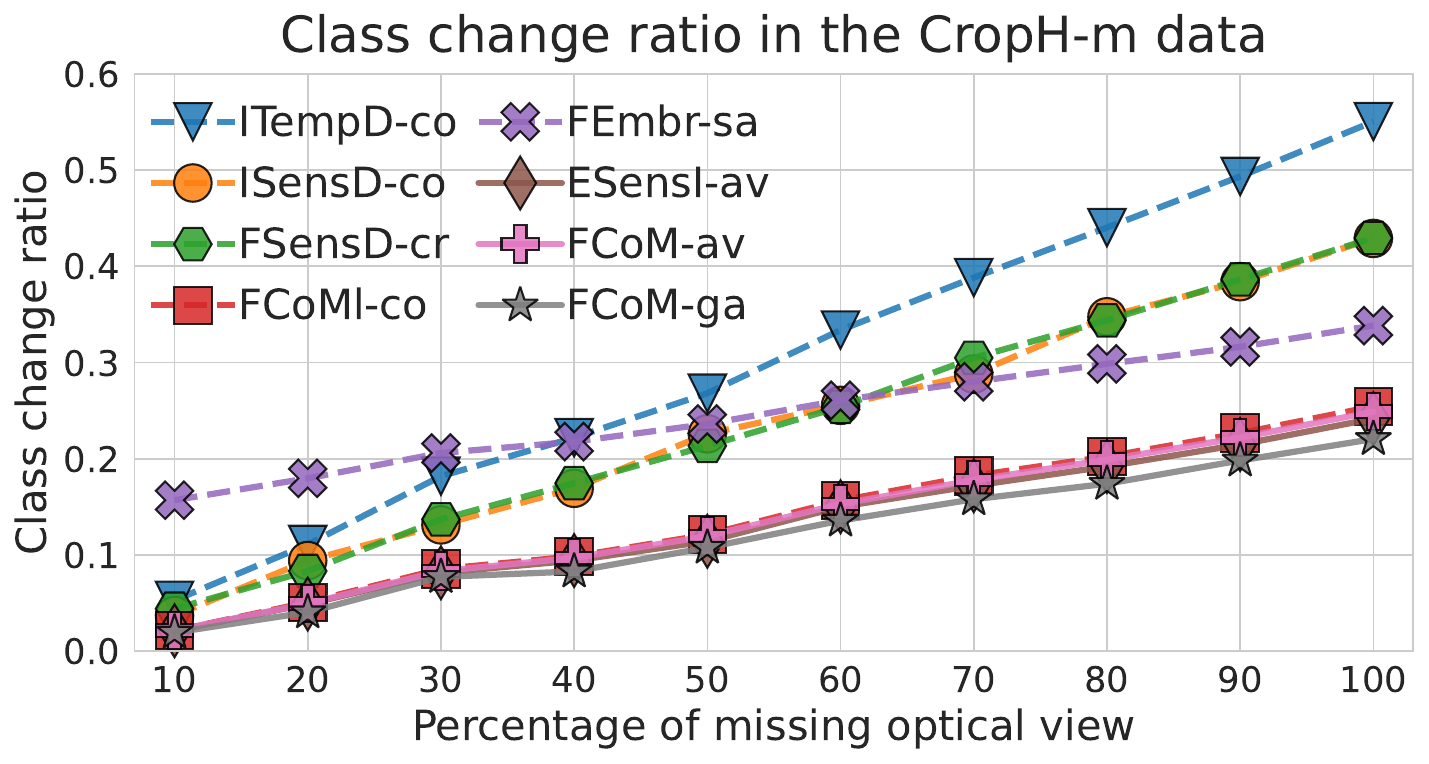}}
\hfill
\subfloat
{\includegraphics[width=0.475\textwidth]{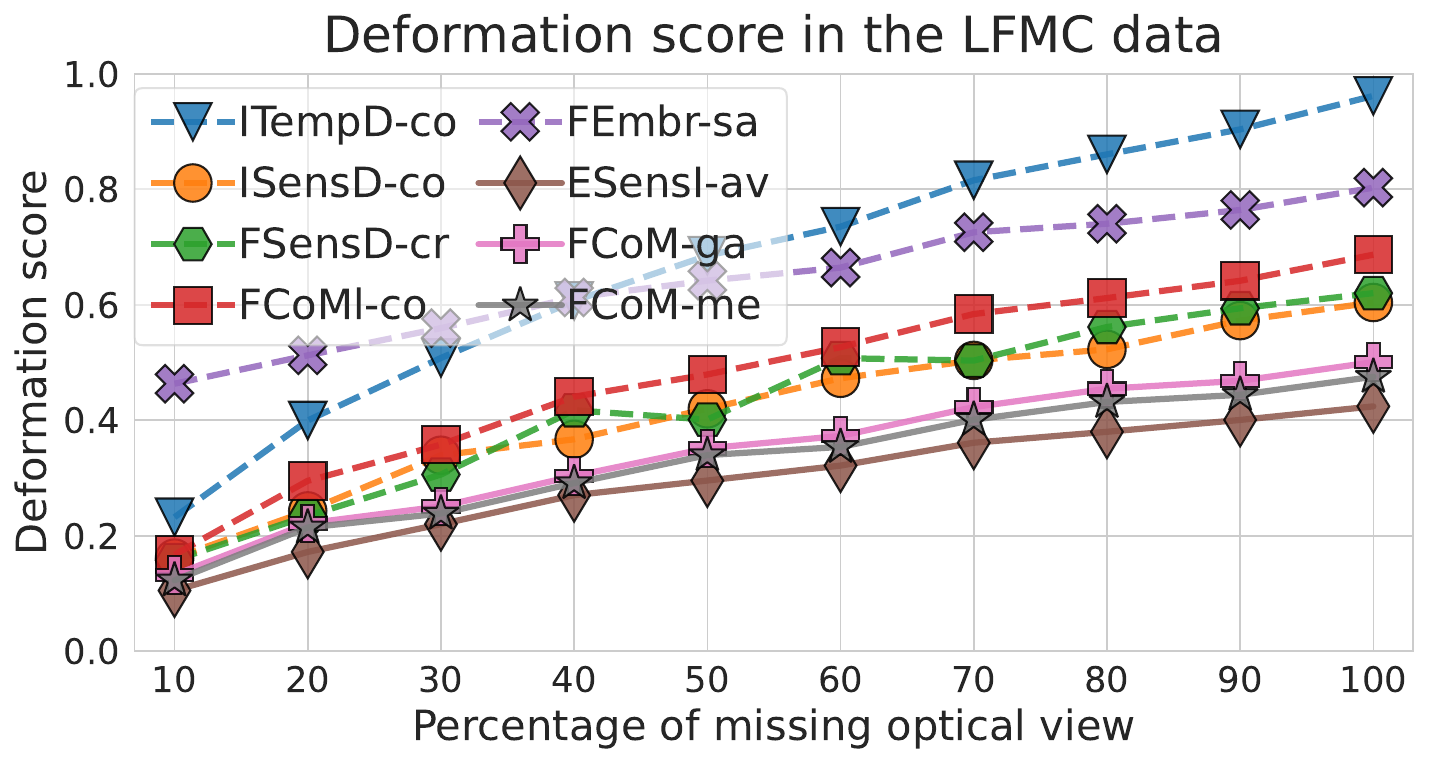}}
\caption{Prediction shift score in classification (class change ratio) and regression (deformation score) tasks.}  \label{fig:perc_missing:shift}
\end{figure}  
We analyze how model predictions are shifted because of missing views, regardless of the target values. 
We plot the class change ratio and the deformation score at different percentages of missing views in the \gls{cropharvestM} and \gls{lfmc} data, respectively, in Figure~\ref{fig:perc_missing:shift}.
The deformation is calculated as the difference in prediction divided by the deviation of the original prediction, i.e. 
$\text{RMSE}(\hat{y}_{\text{full}}, \hat{y}_{\text{miss}})/\text{std}(\hat{y}_{\text{full}})$. 
These graphs show the increase in the prediction shifting when there is more missing data.
We observe that the prediction shift curves of our methods are among the three methods with lower values in the comparison, together with FCoMl-co and ESensI-av methods.

\begin{figure}[t!]
\centering
{\includegraphics[width=0.7\textwidth]{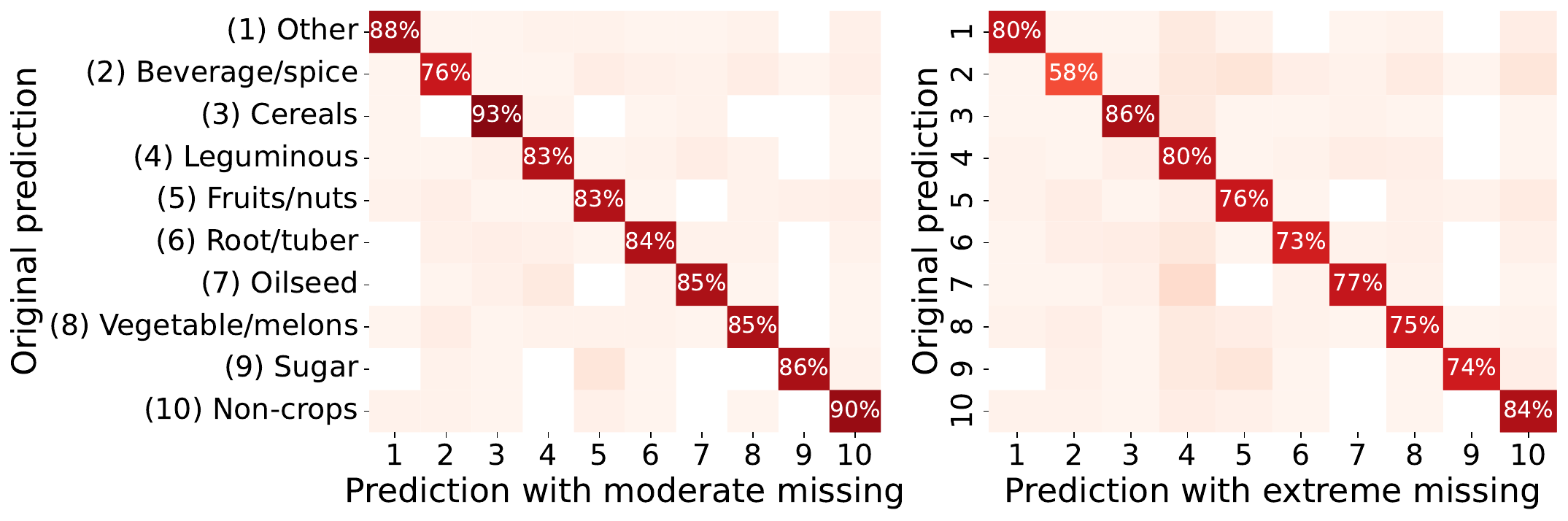}}
\caption{Class prediction shift with moderate (radar view missing), and extreme (optical view available) missingness. The F\gls{allmissing}-ga method is shown in the \gls{cropharvestM} data. The overall class change ratio is $12.5\%$ in the moderate and $20.4\%$ in the extreme cases. 
}\label{fig:shift:cropm}
\end{figure} 
\begin{figure}[t!]
\centering
{\includegraphics[width=0.6\textwidth]{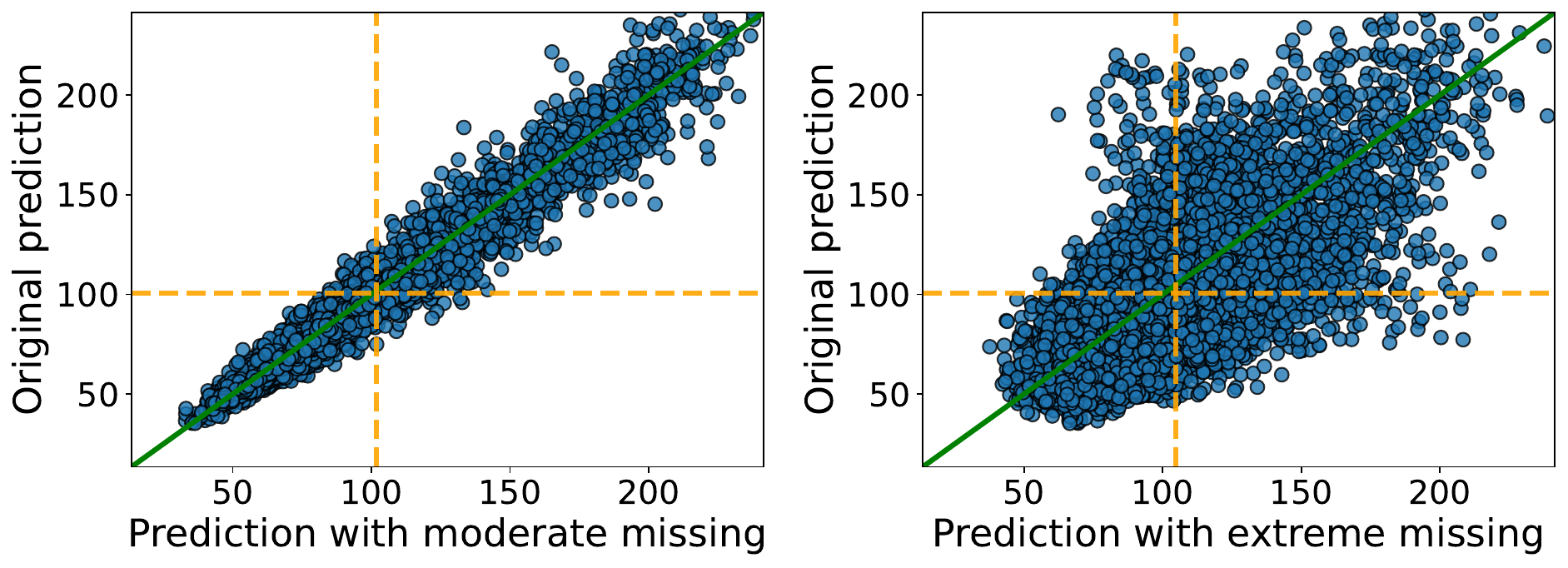}}
\caption{Real-value prediction shift with moderate (radar view missing), and extreme (only optical view available). The F\gls{allmissing}-ga method is shown in the \gls{lfmc} data. The overall deformation score is $0.208$ in moderate and $0.722$ in extreme cases.
}\label{fig:shift:lfmc}
\end{figure} 
As qualitative support for our methods, we plot the class change in the \gls{cropharvestM} data in Figure~\ref{fig:shift:cropm}, and the shift in the real-value predicted in the \gls{lfmc} data in Figure~\ref{fig:shift:lfmc}. 
We notice that in moderate missingness, the prediction change in our methods is insignificant, while in extreme cases is shifted to a greater extent. 
As the predicted value in classification is categorical, the change in prediction is binary (the class change to another one or not), while in the regression task the predicting value is continuous. Therefore, we can see how the prediction is dispersed from the original value to different degrees for all samples.

\subsection{Execution time comparison}

\begin{figure}[t!]
\centering
{\includegraphics[width=0.7\textwidth]{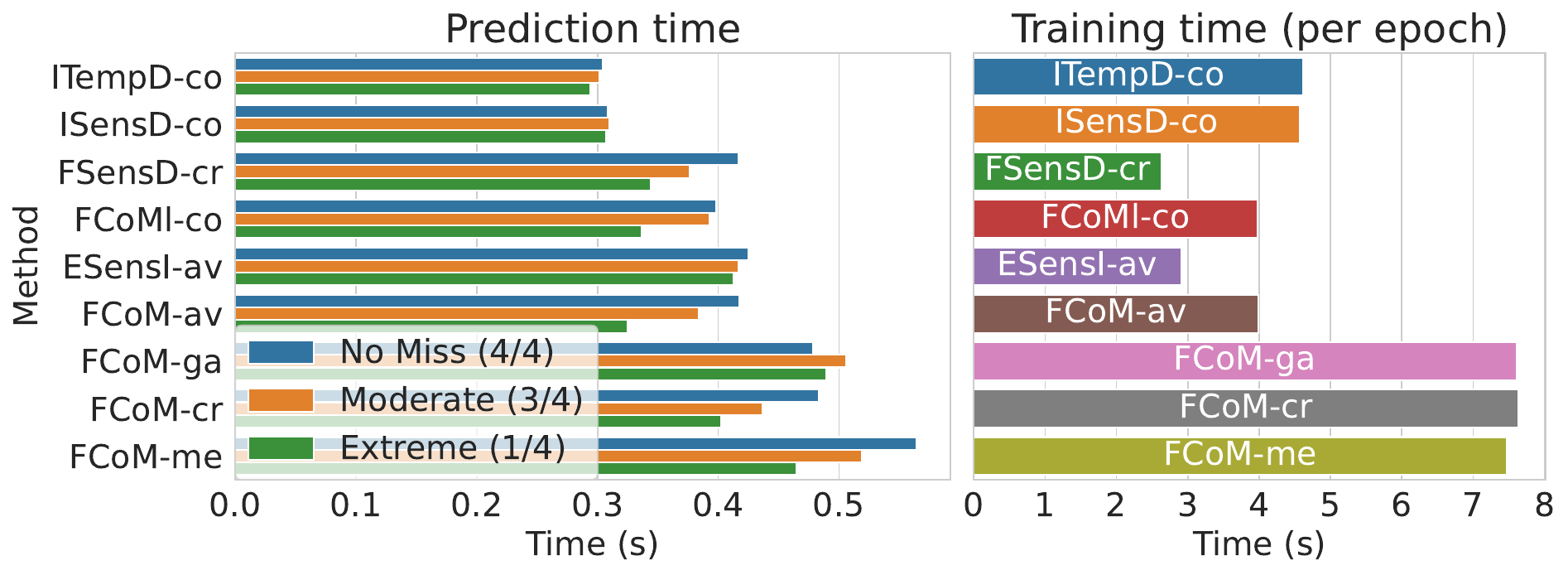}}
\caption{Execution time of different \gls{mvl} methods. The times are calculated in the \gls{cropharvestM} data with 4 views. Prediction times are separated into no missing (4/4), only one view missing (3/4) and only one view available (1/4).}\label{fig:time_comparison}
\end{figure} 
We compare the execution time of different methods in Figure~\ref{fig:time_comparison}. 
During training, we observe that the methods ignoring missing views lead to the most efficient training time, i.e. FSensD-cr, ESensI-av, and FCoM-av. However, we notice that the combination of the \gls{allmissing} technique with sophisticated merge functions (-ga, -cr, and -me suffix) increases training time (per epoch) by almost double.
On the other side, the most efficient prediction time over the entire dataset is associated with the models using input-level fusion. This is expected as these methods use a single encoder architecture. 
Nevertheless, we notice that our methods have a more efficient execution time when there is more missing data, i.e. by ignoring missing views the model dynamical adapts to perform calculations on just the available data (no fake data insertion).

%% file: content/discussion.tex
\section{Analysis and limitation} \label{sec:discussion}

We show the effect of applying \gls{maug}s at different levels of the \gls{mvl} models with average fusion in Table~\ref{tab_abl:maug}. 
Similar results are obtained with other merge functions (see \ref{sec_app:merge_comp}).
We zero-impute views if they are missing at input-level, and ignore them if missing at feature-level.
We notice a trend to increase the model robustness to missing views when the \gls{allmissing} is used, compared to \gls{sensd} and without \gls{maug}s. 
Randomly dropping sensors, used in \citet{chen2024novel,mena2024notyet}, have good relative robustness but quite low overall performance.
In the \gls{lfmc} data, we find it more effective to address missing views at the input rather than feature level, as noted in the literature \cite{mena2024igarss}.
However, with the \gls{allmissing} technique, we notice that is better to use it at feature by ignoring than at input-level with zero imputation.
Besides, the \gls{allmissing} even allows increasing the full-view performance in the \gls{cropharvestM} data.
The evidence suggests that, in most cases, our usage of \gls{allmissing} at feature-level and dynamic fusion is optimal for model robustness.

\begin{table*}[!t]
\caption{Different \gls{maug} techniques applied at input (concatenation) and feature (average) level. The $\dagger$ is a value below -10.} \label{tab_abl:maug} 
\scriptsize
\begin{tabularx}{\textwidth}{ll|CCCCC|CCCCC} \cmidrule{3-12}
        \multicolumn{2}{c}{} & \multicolumn{5}{c}{\textbf{\gls{cropharvestM}} (\gls{f1} scores)} & \multicolumn{5}{c}{\textbf{\gls{lfmc}} (\gls{r2} scores)} \\
        \toprule
        & & (4/4) No & \multicolumn{2}{c}{(3/4) Only missing} & \multicolumn{2}{c|}{(1/4) Only available} & (6/6) No & \multicolumn{2}{c}{(5/6) Only missing} & \multicolumn{2}{c}{(1/6) Only available} \\
        \gls{maug} & Level & Missing & \multicolumn{1}{c}{Radar}  & Optical  & \multicolumn{1}{c}{Optical} & Radar  &  Missing & \multicolumn{1}{c}{Radar}  & Optical  & \multicolumn{1}{c}{Optical} & Radar 
        \\    
        \midrule
         - & Input & $\secondbest{0.655}$ & $0.569$ & $0.359$ & $0.230$ & $0.081$ &  $\secondbest{0.638}$	& $\secondbest{0.593}$ & $0.262$ & ${0.293}$ & ${0.057}$ \\
         \gls{sensd} & Input & 	$0.566$ &	$0.522$ &	$0.422$ &	$0.401$ &	$0.206$ & $0.506$ &	$0.468$ &	$0.342$ &	$\secondbest{0.305}$ &	$\secondbest{0.134}$ \\
         \gls{allmissing} & Input &  $0.652$ & $\secondbest{0.637}$ & $0.385$ & $\secondbest{0.602}$ & $0.123$  &  $0.515$	& $0.484$ & $\secondbest{0.355}$ & $\best{0.317}$ & $\best{0.118}$\\  
         - & Feature &  $0.647$ & $0.561$ & ${0.492}$ & ${0.269}$ & ${0.224}$ & $\best{0.648}$ & $0.531$ & $0.151$ & $\dagger$ & $\dagger$ \\
         \gls{sensd} & Feature &  $0.634$ &	$0.615$ &	$\secondbest{0.526}$ &	$0.590$ &	$\secondbest{0.386}$ &  $0.511$ &	$0.496$ &	$0.302$ &	$\dagger$ &	-$1.709$  \\ 
         \gls{allmissing} & Feature & $\best{0.679}$ & $\best{0.646}$ & $\best{0.549}$ & $\best{0.645}$ & $\best{0.430}$ & ${0.625}$ & $\best{0.604}$ & $\best{0.437}$ & -$9.632$ & -$8.362$  \\
         \bottomrule
    \end{tabularx}
\end{table*}

Along the results, we note a slight variation in which method obtains the best results for each dataset and missing view scenario. However, this variability is expected in the \gls{eo} field, as the data is quite heterogeneous and region-dependent \cite{camps2021deep,hong2021more,ofori-ampofo2021crop,mena2023comparative,rolf2024mission}. 
In addition, this variation depends on the metrics used to assess the models. In our study, standard performance metrics are used from the literature. However, additional values are included in the \ref{sec_app:add_metrics}.
Nevertheless, our combination of \gls{allmissing} with the simple average function (F\gls{allmissing}-av) shows good overall results, without significantly increasing the training time, as well as having an adaptive prediction time based on the available views.

We remark that Transformer models are not naturally robust to missing data \cite{ma2022multimodal}. They can handle missing data without intervention, but that does not imply they will obtain the same performance as when there is no missing data \cite{du2023saits,chen2024novel}. For instance, in our case, we show that by using the cross-attention fusion (based on Transformer models) is not optimal for our problem. 

Regarding the limitations, we state that the \gls{allmissing} technique is dependent on the total views of each dataset ($m$), since the number of augmented samples in the \gls{allmissing} technique depends on this: $2^{m} -1$.
In addition, we assess the effect of missing views only at inference, assuming a full-view training dataset.
However, based on the dynamic fusion, it could be easily extended to other settings.
Furthermore, we validate our methods using pixel-wise \gls{eo} datasets. Although this validation is conducted across four datasets, its effectiveness in other domains needs to be verified.

%% file: content/conclusion.tex
\section{Conclusion} \label{sec:conclusion}

We introduce \acrfull{maug} technique tailored to \acrfull{mvl} with missing views, named \acrfull{allmissing}. 
We apply this technique at feature-level in combination with dynamic merge functions. 
For evaluation, we simulate missing views during inference to assess model robustness (predictive performance decay) in two scenarios. Moderate missingness, when only \textit{top} views (the ones with the best individual performance) are missing, and extreme missingness, when only one \textit{top} view is available.
Our findings show that our approaches outperform competing methods in moderate missingness, particularly with F\gls{allmissing}-av. Moreover, due to the \gls{maug} effect, we see an increase in the classification performance in the full-view scenario.
In addition, we identify challenging scenarios for robustness, particularly in regression tasks and extreme missingness. 
For these challenges, future work should consider developing models that operate with any available data at decision-level fusion, such as by weighing predictions with missing data.

%% file: content/supp.tex
\section{Initial setup}

\subsection{Dataset Description} \label{sec_app:data}

We show the features from each view in the different datasets in Tables \ref{tab_supp:data:pm25}, \ref{tab_supp:data:crop}, and \ref{tab_supp:data:lfmc} for \gls{pm25}, CropHarvest, and \gls{lfmc} data respectively.
The abbreviations used in the tables correspond to: normalized difference vegetation index (NDVI), normalized difference water index (NDWI), and near infrared vegetation index (NIRv).
\begin{table*}[h!]
    \caption{Name of features in each view in the \textbf{\gls{pm25}} data \cite{pm25}. The source of the views are ground-based stations.} \label{tab_supp:data:pm25}
    \centering
    \footnotesize
    \begin{tabular}{l|l}
    \toprule
         View & Features \\ 
         \midrule
         Conditions & dew point, temperature, and humidity \\
         Dynamics & pressure, combined wind direction, cumulated wind speed, season \\
         Precipitation & precipitation, and cumulated precipitation \\
         \bottomrule
    \end{tabular}
\end{table*}
\begin{table*}[h!]
    \caption{Name of features in each view in the \textbf{\gls{cropharvestB}} and \textbf{\gls{cropharvestM}} data \cite{tseng2021crop}. The views are at 10 m spatial resolution.} \label{tab_supp:data:crop}
    \centering
    \footnotesize
    \begin{tabular}{l|l|l} \toprule
         View & Source & Features \\ 
         \midrule
         Optical & Sentinel-2 (level 1C) & B2 (blue), B3 (green), B4 (red), B5, B6, B7, B8, B8A, B9, B11, B12, NDVI \\
         Radar &  Sentinel-1 (C-band) & VV and VH polarization bands \\
         Weather & ERA5 & temperature and precipitation \\
         Topographic & NASA's SRTM & elevation and slope \\
         \bottomrule
    \end{tabular}
\end{table*}
\begin{table*}[h!]
    \caption{Name of features in each view in the \textbf{\gls{lfmc}} data \cite{rao2020sar}. The views are at 250 m spatial resolution.} \label{tab_supp:data:lfmc}
    \centering
    \footnotesize
    \begin{tabular}{l|l|l} \toprule
         View & Source & Features \\ 
         \midrule
         Optical & Landsat 8 & red, green, blue, near infrared, short-wave infrared, NDVI, NDWI, NIRv \\
         Radar &  Sentinel-1 (C-Band) & VV, VH, and VH/VV polarization bands \\
         Topographic & National Elevation Database & elevation and slope \\
         Soil & Unified North American Soil Map & silt, sand, and clay \\
         LiDAR & Global Laser Altimetry System & canopy height (ordinal value) \\
         Land-cover & GLOBCOVER & class label between 12 options \\
         \bottomrule
    \end{tabular}
\end{table*}

\subsection{Architecture Selection} \label{sec_app:arch_sel}

In Table~\ref{tab_abl:rnn_permu} we compare the \gls{maug} techniques to a view-permutation in the memory-based fusion with a LSTM architecture. 
Since views are processed sequentially, the view-permutation is included to prevent the model from overfitting the views order \cite{lee2019set}. 
We observe that the usage of the \gls{maug} technique has a greater positive effect on the predictive performance than the view-permutation in the full-view scenario and with missing views. 
This suggests that the memory fusion with the \gls{maug} does not require an explicit permutation to become order invariant and increase generalization.
Nevertheless, the good behavior without permutation in \gls{lfmc} data might be just overfitting, caused by the small dataset (less than 2000 samples to train).

\begin{table}[!h]
    \caption{Memory fusion-based \gls{mvl} model with different configurations of view-permutation and \gls{maug} techniques.} \label{tab_abl:rnn_permu}
    \centering
    \footnotesize
    \begin{tabular}{ll|cc|cc} \cmidrule{3-6}
          \multicolumn{2}{c}{}  &  \multicolumn{2}{c}{\textbf{\gls{cropharvestM}} (\gls{f1})} &  \multicolumn{2}{c}{\textbf{\gls{lfmc}} (\gls{r2})} \\
          \toprule
         \gls{maug} & Permutation & No Missing & Missing Optical & No Missing & Missing Optical \\
         \midrule
         - & - &   $0.656$  & $0.503$ & $\best{0.735}$ & $0.095$	 \\
         - &  All & $0.649$ & $0.487$ & $\secondbest{0.706}$ & $0.192$ \\
         \gls{sensd} & All & $0.643$ & $0.536$ &  $0.551$ & $0.353$ \\
         - & Random &  $0.643$ & $0.496$	 & $0.666$ & $0.224$ \\ 
         \gls{sensd} & Random &   $0.641$ & $0.531$ & $0.507$ & $0.327$ \\
         \gls{allmissing} & Random & $\secondbest{0.666}$ & $\secondbest{0.555}$ &  $0.595$ & $\secondbest{0.428}$	 \\
         \gls{allmissing} & - &  $\best{0.672}$ & $\best{0.559}$ & $0.613$  & 	$\best{0.412}$	\\
         \bottomrule
    \end{tabular}
\end{table}

In Table~\ref{tab_abl:rnn_arch} we compare different architectures of the memory fusion.
We notice a tendency to get better results (in full-view and with missing views) with a more complex network architecture. 
Overall, the best results are obtained with two bidirectional LSTM layers, while the second best results are associated with an architecture based on GRU layers.
\begin{table}[!t]
    \caption{Memory fusion-based \gls{mvl} model with different network architectures. ``Bi'' stands for bidirectional layer.} \label{tab_abl:rnn_arch}
    \centering
    \footnotesize
    \begin{tabular}{ll|cc|cc} \cmidrule{3-6}
          \multicolumn{2}{c}{}  &  \multicolumn{2}{c}{\textbf{\gls{cropharvestM}} (\gls{f1})} &  \multicolumn{2}{c}{\textbf{\gls{lfmc}} (\gls{r2})} \\
          \toprule
         Gate & Layers & No Missing & Missing Optical & No Missing & Missing Optical \\
         \midrule
         GRU & 1 & $0.666$ & $\secondbest{0.554}$ & $0.627$ & $0.422$ \\
         GRU & 1 (Bi) & $\secondbest{0.671}$ & $\secondbest{0.554}$ & $\secondbest{0.636}$ & $0.451$ \\
         GRU & 2 (Bi) & $0.663$ & $0.552$ & $0.634$ & $\secondbest{0.468}$ \\
         GRU & 3 (Bi) & $0.661$ & $0.546$ & $0.626$ & $0.450$ \\
         LSTM & 1 (Bi) & $0.670$ & $0.552$ & $0.634$ & $0.461$ \\
         LSTM & 2 (Bi) & $\best{0.677}$ & $\best{0.557}$ & $\best{0.657}$ & $\best{0.477}$ \\
         \bottomrule
    \end{tabular}
\end{table}

In Table~\ref{tab_abl:cross_arch} we compare different architectural options of the cross-attention fusion, based on Transformer layers.
We observe an optimal value across datasets and missing scenarios with eight heads and just one layer.
Besides, we notice a slight tendency to get better results when using a more complex network architecture.
\begin{table}[!t]
    \caption{Cross-attention fusion-based \gls{mvl} model with different network architectures.} \label{tab_abl:cross_arch}
    \centering
    \footnotesize
    \begin{tabular}{cc|cc|cc} \cmidrule{3-6}
          \multicolumn{2}{c}{}  &  \multicolumn{2}{c}{\textbf{\gls{cropharvestM}} (\gls{f1})} &  \multicolumn{2}{c}{\textbf{\gls{lfmc}} (\gls{r2})} \\
          \toprule
         Heads & Layers & No missing & Missing optical & No missing & Missing optical \\
         \midrule
         1 & 1 & $0.637$ & $0.468$ & $0.545$ & $0.375$ \\
         2 & 1 & $0.622$ & $0.462$ & ${0.551}$ & $\secondbest{0.380}$ \\
         4 & 1 & $0.650$ & $0.511$ & $\secondbest{0.553}$ & $\secondbest{0.380}$ \\
         4 & 2 & $\secondbest{0.654}$ & $0.511$ & $0.536$ & $0.351$ \\
         8 & 1 & $\best{0.659}$ & $\best{0.520}$ & $\best{0.569}$ & $\best{0.400}$ \\
         8 & 2 & $0.650$ & $0.504$ & $0.541$ & $0.371$ \\
         8 & 3 & $0.652$ & $\secondbest{0.519}$ & $0.541$ & $0.362$ \\
         \bottomrule
    \end{tabular}
\end{table}

\subsection{Individual view performance} \label{sec_app:ind}
\begin{table}[b!]
\caption{Predictive performance of individually trained models (per view). The \gls{f1} scores are shown in classification tasks and \gls{r2} scores in the regression tasks. The \textbest{best} and \textsecondbest{second best} values are highlighted. }\label{tab:individual}
\centering
\footnotesize
\begin{tabular}{l|cccc}
\toprule
View & \textbf{\gls{cropharvestB}} & \textbf{\gls{cropharvestM}} & \textbf{\gls{lfmc}} & \textbf{\gls{pm25}} \\
\midrule
Optical & $\best{0.791} \std{0.013}$  & $\best{0.635} \std{0.023}	$ & $\best{0.194} \std{0.224}$ &  \\
Radar &  $\secondbest{0.752} \std{0.012}	$ & $\secondbest{0.444} \std{0.022}$ & $\secondbest{0.050} \std{0.360}$ & \\
Topographic & $0.631 \std{0.044}$	& $0.095 \std{0.028}$ & -$0.124 \std{0.590}$ & \\
Weather & $0.701 \std{0.012}$ & $0.346 \std{0.013}$ &  &  \\
Soil &  & & -$0.245 	\std{0.557}$ & \\
LiDAR &  & & -$0.033 \std{0.147}$ & \\
Land-cover  & & &  -$0.021 \std{0.102}$ &  \\
Conditions & & & & $\secondbest{0.034} \std{0.135}$ \\
Dynamics & & & & $\best{0.334} 	\std{0.078}$  \\
Precipitation & & & & -$0.072 \std{0.068}$ \\
\bottomrule
\end{tabular}
\end{table}

In order to detect the \textit{top} views for prediction in each dataset, we train an individual model on each view. The results for each dataset are shown in Table~\ref{tab:individual}.
For \gls{cropharvestB}, \gls{cropharvestM}, and \gls{lfmc} these are optical and radar views, while for the \gls{pm25} dataset the top views are dynamic and condition. 
We note that the static views usually have a low predictive performance, only serving as a complement to the \textit{top} views for prediction.

\section{Additional results}

\subsection{Another top view missing} \label{sec_app:another_view}

\begin{figure*}[h!]
\centering
\subfloat
{\includegraphics[width=0.475\linewidth]{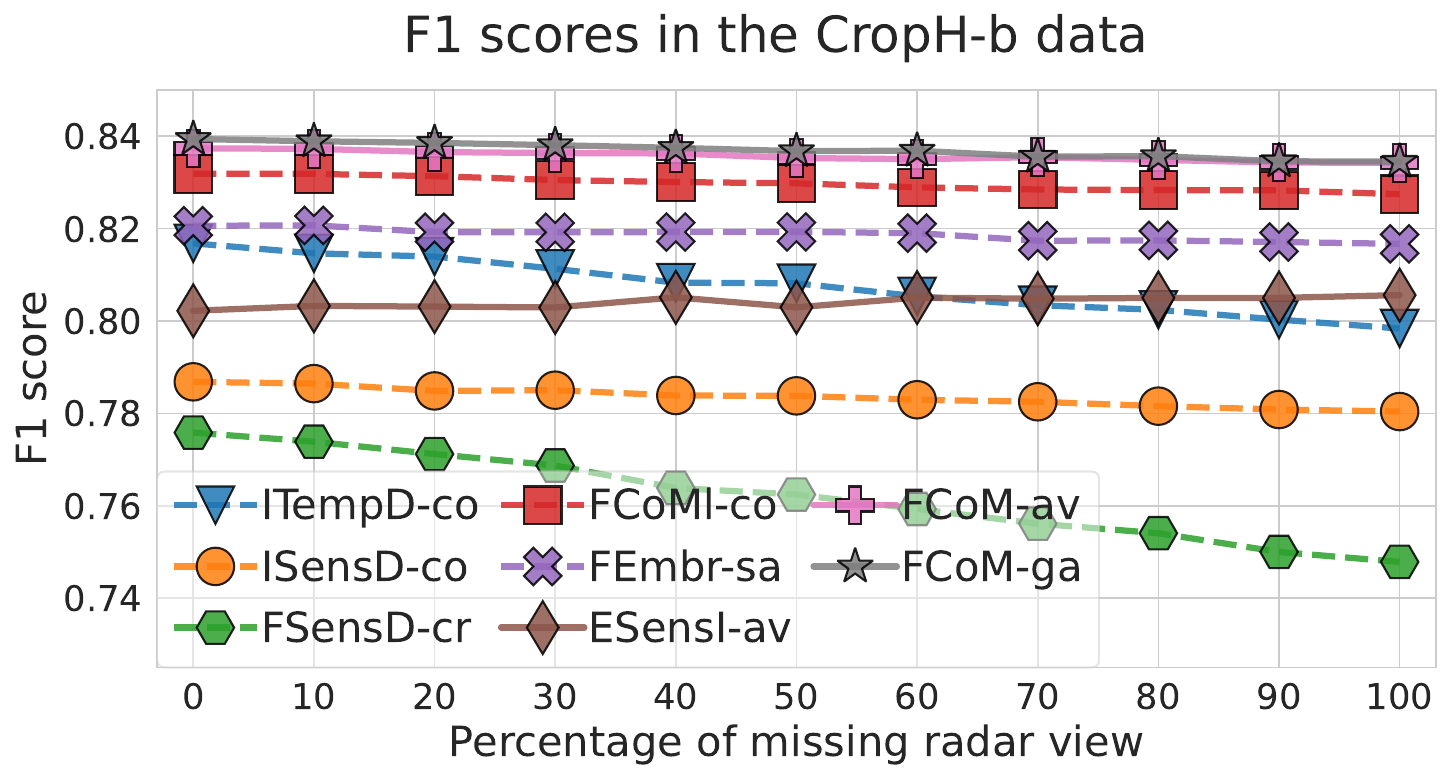}}
\hfill
{\includegraphics[width=0.475\linewidth]{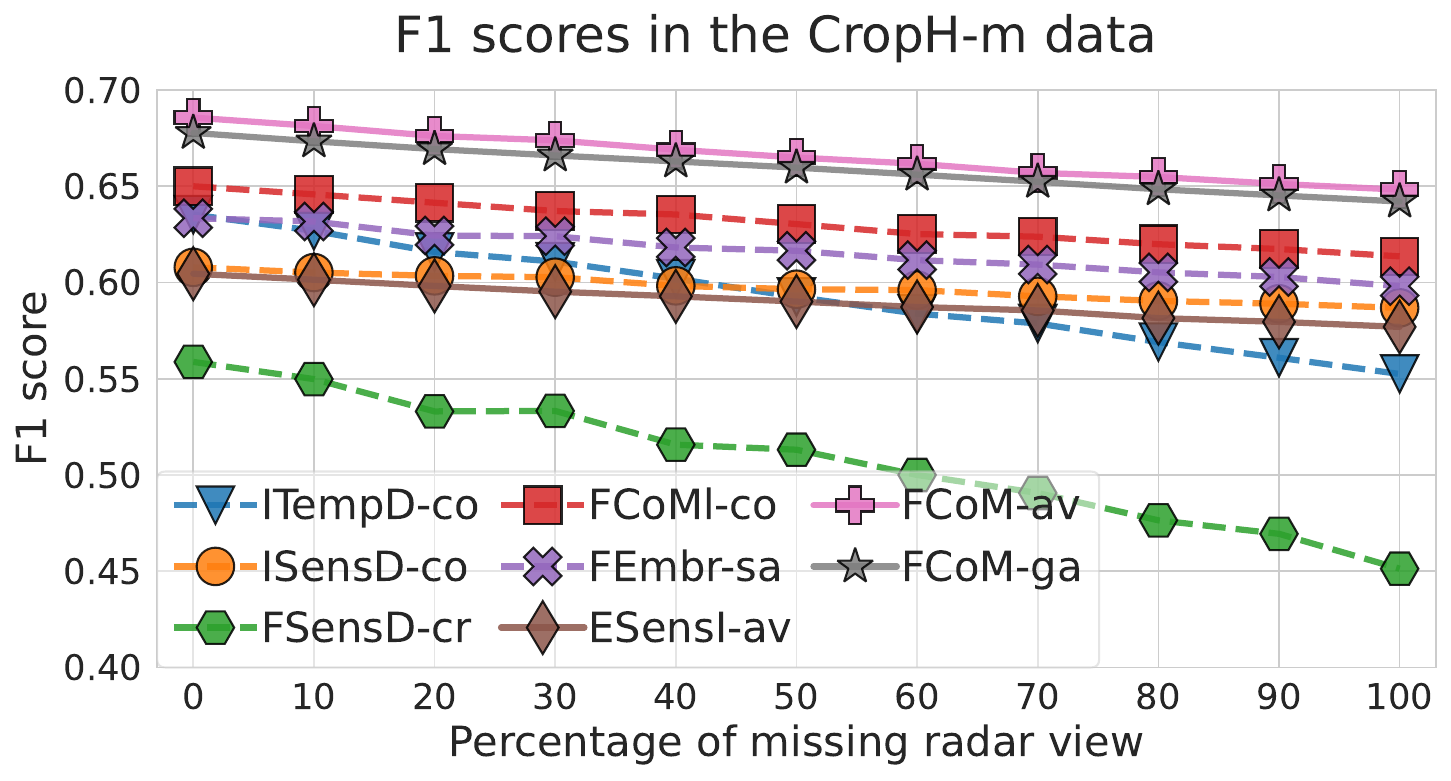}}
\\
\subfloat
{\includegraphics[width=0.475\linewidth]{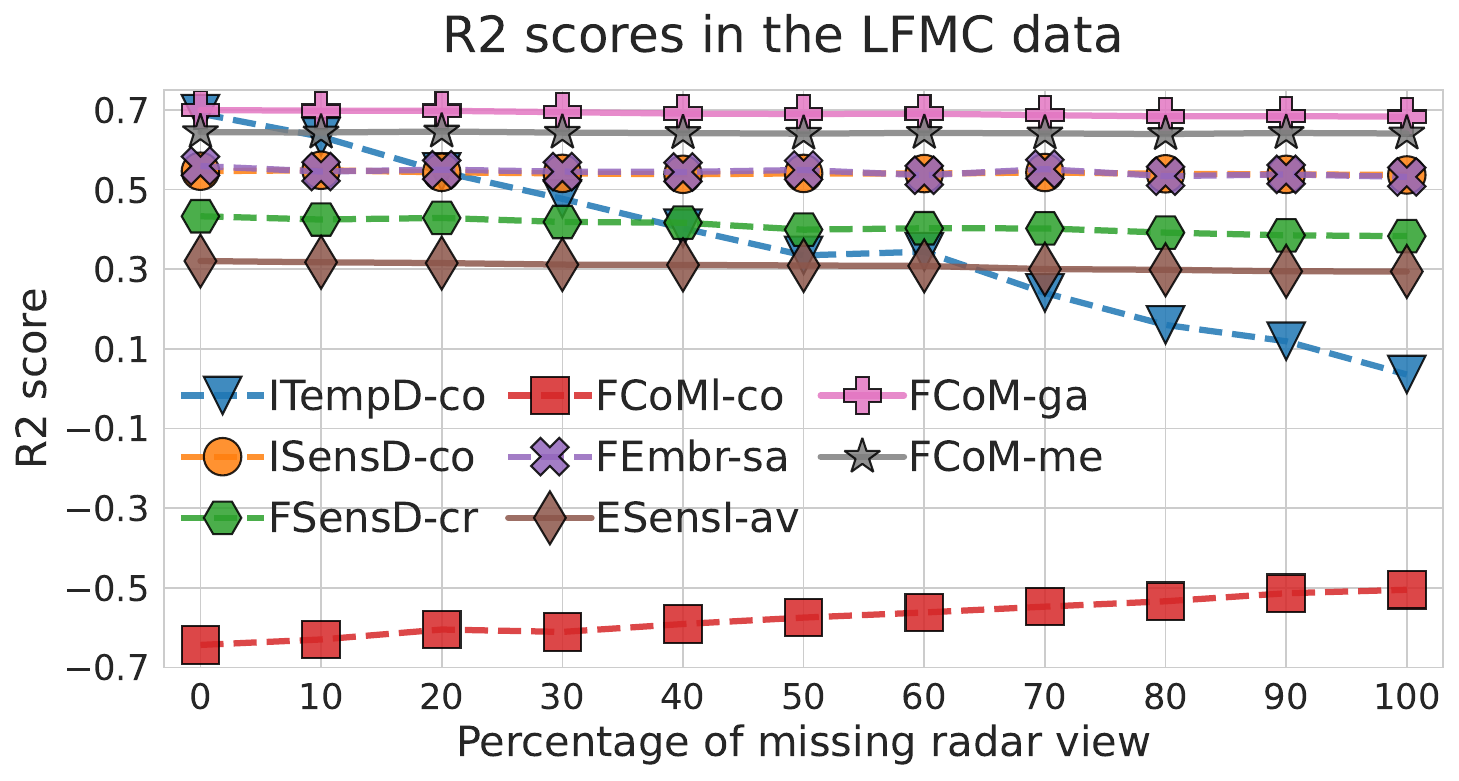}}
\hfill
\subfloat
{\includegraphics[width=0.475\linewidth]{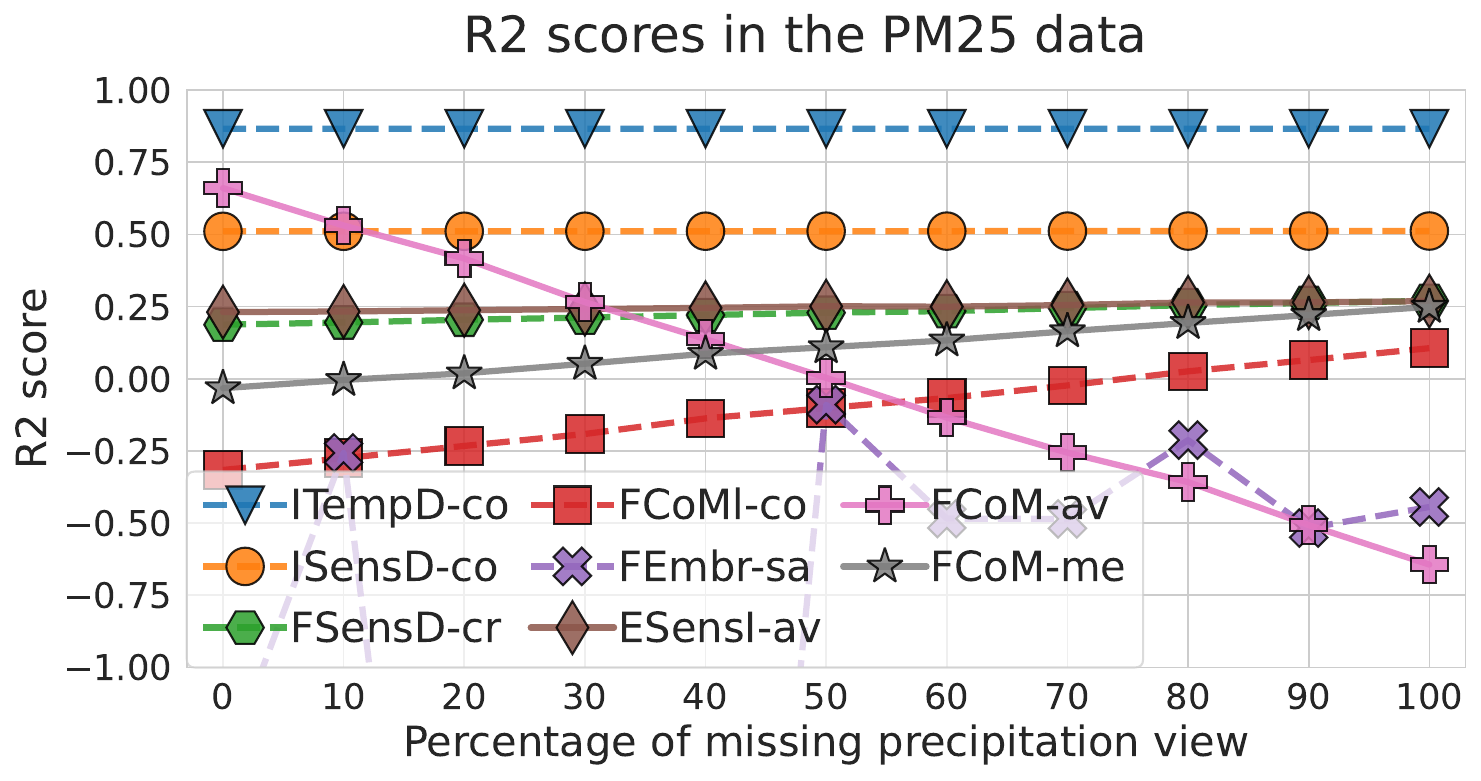}}
\caption{Predictive performance and robustness when varying percentages of validation samples have a \textit{top} view missing.}\label{fig_supp:perc_missing_add}
\end{figure*}

In Figure \ref{fig_supp:perc_missing_add} we display the predictive performance in all datasets when additional views are missing in some samples, as a complement to Figure~\ref{fig:perc_missing}. We present our two best methods in each dataset.
In the classification tasks, the proposed F\gls{allmissing}-ga method has the best predictive performance along the percentage of missing data, as observed when the top view is missing (Figure~\ref{fig:perc_missing}).
Overall, our methods show the best behavior (of a good balance between a small slope and a high value) when increasing the level of samples with missing views. 
Furthermore, most of the methods have a high robustness to missing the radar view in the classification tasks.
Surprisingly, F\gls{allmissing}-av has a strange behavior in the \gls{pm25} data, being the only one greatly affected by missing the precipitation view. Perhaps our method learned a prediction quite dependent on this view in that dataset. 

\subsection{Dynamic merge function comparison}\label{sec_app:merge_comp}

\begin{table*}[!b] 
        \caption{Different configurations of the \gls{maug} technique applied at input or feature level with \textbf{gated fusion}. The \gls{f1} scores are shown in the \gls{cropharvestM} data and \gls{r2} scores in the \gls{lfmc} data. The $\dagger$ is a value below -10.} \label{tab_supp:maug_gated}
    \centering
    \scriptsize 
    \begin{tabularx}{\textwidth}{ll|CCCCC|CCCCC} \cmidrule{3-12}
        \multicolumn{2}{c}{} & \multicolumn{5}{c|}{\textbf{\gls{cropharvestM}} (\gls{f1})} & \multicolumn{5}{c}{\textbf{\gls{lfmc}} (\gls{r2})} \\ 
        \toprule
        & & (4/4) No & \multicolumn{2}{|c}{Only missing (3/4)} & \multicolumn{2}{|c|}{Only available (1/4)} & (6/6) No & \multicolumn{2}{|c}{Only missing (5/6)} & \multicolumn{2}{|c}{Only available (1/6)} \\
        \gls{maug} & Level & missing  & \multicolumn{1}{|c}{Radar}  & Optical  & \multicolumn{1}{|c}{Optical} & Radar  &  missing & \multicolumn{1}{|c}{Radar}  & Optical  & \multicolumn{1}{|c}{Optical} & Radar 
        \\ 
         \midrule
         - & Input &$0.643$ &	$0.506$ &	$0.354$ &	$0.189$ &	$0.073$ & $\secondbest{0.735}$ &	$0.631$ &	$0.238$ &	$0.215$ &	-$0.011$   \\
         \gls{sensd} & Input & $0.592$ &	$0.562$ &	$0.479$ &	$0.505$ &	$0.323$ & $0.545$ &	$0.522$ &	$0.324$ &	-$3.148$ &	-$5.625$ \\
         \gls{allmissing} & Input &  $0.662$ &	$0.619$ &	$0.417$ &	$0.565$ &	$0.089$ & $0.615$ &	$0.620$ &	$0.292$ &	$\secondbest{0.234}$ &	$\secondbest{0.012}$ \\  
         - & Feature &  $0.653$ &	$0.574$ &	$0.494$ &	$0.355$ &	$0.250$ & $\best{0.738}$ &	$\secondbest{0.660}$ &	$0.225$ &	$\dagger$ &	$\dagger$ \\
         \gls{sensd} & Feature & $\secondbest{0.668}$ &	$\secondbest{0.632}$ &	$\secondbest{0.538}$ &	$\secondbest{0.610}$ &	$\secondbest{0.387}$  & $0.532$ &	$0.499$ &	$\secondbest{0.336}$ &	-$3.151$ &	-$0.550$ \\ 
         \gls{allmissing} & Feature & $\best{0.678}$ &	$\best{0.647}$ &	$\best{0.568}$ &	$\best{0.627}$ &	$\best{0.418}$ & $0.684$ &	$\best{0.678}$ &	$\best{0.471}$ &	$\best{0.326}$ &	$\best{0.158}$  \\
         \bottomrule
    \end{tabularx}
\end{table*}

\begin{table*}[!t]
        \caption{Different configurations of the \gls{maug} technique applied at input or feature level with \textbf{cross-attention fusion}. The \gls{f1} scores are shown in the \gls{cropharvestM} data and \gls{r2} scores in the \gls{lfmc} data. The $\dagger$ is a value below -10.} \label{tab_supp:maug_cross}
        \centering
        \scriptsize
    \begin{tabularx}{\textwidth}{ll|CCCCC|CCCCC} \cmidrule{3-12}
        \multicolumn{2}{c}{} & \multicolumn{5}{c|}{\textbf{\gls{cropharvestM}} (\gls{f1})} & \multicolumn{5}{c}{\textbf{\gls{lfmc}} (\gls{r2})} \\ 
        \toprule
        & & (4/4) No & \multicolumn{2}{|c}{Only missing (3/4)} & \multicolumn{2}{|c|}{Only available (1/4)} & (6/6) No & \multicolumn{2}{|c}{Only missing (5/6)} & \multicolumn{2}{|c}{Only available (1/6)} \\
        \gls{maug} & Level & missing & \multicolumn{1}{|c}{Radar}  & Optical  & \multicolumn{1}{|c}{Optical} & Radar  & missing  & \multicolumn{1}{|c}{Radar}  & Optical  & \multicolumn{1}{|c}{Optical} & Radar 
        \\ 
         \midrule
         - & Input & $0.646$ &	$0.585$ &	$0.239$ &	$0.432$ &	$0.072$ & $\secondbest{0.575}$ &	$0.521$ &	$0.289$ &	$\secondbest{0.242}$ &	$\secondbest{0.053}$  \\
         \gls{sensd} & Input & $0.569$ &	$0.54$ &	$0.438$ &	$0.523$ &	$0.302$ & $0.522$ &	$0.503$ &	${0.303}$ &	-$0.005$ &	-$0.205$ \\
         \gls{allmissing} & Input & $\secondbest{0.655}$ &	$0.636$ &	$0.216$ &	$0.620$ &	$0.071$ & 	$0.554$ &	$0.481$ &	$\best{0.361}$ &	$\best{0.295}$ &	$\best{0.089}$ \\  
         - & Feature &  	$0.637$ &	$0.589$ &	$0.495$ &	$0.506$ &	$0.287$ & $\best{0.579}$ &	$\best{0.539}$ &	-$0.388$ &	-$0.058$ &	-$2.702$ \\
         \gls{sensd} & Feature & $0.639$ &	$\secondbest{0.616}$ &	$\secondbest{0.510}$ &	$\secondbest{0.617}$ &	$\secondbest{0.403}$ & 	$0.536$ &	$\secondbest{0.525}$ &	$\secondbest{0.315}$ &	$0.013$ &	$0.019$ \\ 
         \gls{allmissing} & Feature & $\best{0.665}$ &	$\best{0.636}$ &	$\best{0.547}$ &	$\best{0.638}$ &	$\best{0.441}$ & $0.520$ &	$0.471$ &	${0.278}$ &	-$0.363$ &	-$1.005$ \\
         \bottomrule
    \end{tabularx}
\end{table*}
\begin{table*}[!t]
        \caption{Different configurations of the \gls{maug} technique applied at input or feature level with \textbf{memory fusion}. The \gls{f1} scores are shown in the \gls{cropharvestM} data and \gls{r2} scores are shown in the \gls{lfmc} data.} \label{tab_supp:maug_mem}
        \centering
    \scriptsize
    \begin{tabularx}{\textwidth}{ll|CCCCC|CCCCC} \cmidrule{3-12}
        \multicolumn{2}{c}{} & \multicolumn{5}{c|}{\textbf{\gls{cropharvestM}} (\gls{f1})} & \multicolumn{5}{c}{\textbf{\gls{lfmc}} (\gls{r2})} \\ 
        \toprule
        & & (4/4) No & \multicolumn{2}{|c}{Only missing (3/4)} & \multicolumn{2}{|c|}{Only available (1/4)} & (6/6) No & \multicolumn{2}{|c}{Only missing (5/6)} & \multicolumn{2}{|c}{Only available (1/6)} \\
        \gls{maug} & Level & missing & \multicolumn{1}{|c}{Radar}  & Optical  & \multicolumn{1}{|c}{Optical} & Radar  & missing & \multicolumn{1}{|c}{Radar}  & Optical  & \multicolumn{1}{|c}{Optical} & Radar 
        \\ 
         \midrule
         - & Input & ${0.651}$ &	$0.572$ &	$0.492$ &	$0.287$ &	$0.237$ & $\secondbest{0.735}$ &	$\secondbest{0.634}$ &	$0.072$ &	-$4.165$ &	$\dagger$ \\
         \gls{sensd} & Input &  $0.586$ &	$0.547$ &	$0.471$ &	$0.490$ &	$0.319$ &  $0.568$ &	$0.549$ &	$0.359$ &	-$0.015$ &	-$0.383$ \\
         \gls{allmissing} & Input & $\best{0.662}$ &	$\best{0.630}$ &	$\secondbest{0.556}$ &	$\best{0.618}$ &	$\secondbest{0.413}$  &  $0.648$ &	$0.630$ &	$\best{0.420}$ &	$\best{0.346}$ &	$\secondbest{0.146}$ \\  
         - & Feature & $0.641$ &	$0.568$ &	$0.499$ &	$0.316$ &	$0.235$ & $\best{0.741}$ &	$\best{0.671}$ &	$0.022$ &	-$0.753$ &	-$3.337$ \\
         \gls{sensd} & Feature & $0.638$ &	${0.610}$ &	${0.533}$ &	$0.594$ &	$0.378$ & $0.557$ &	$0.547$ &	$0.372$ &	-$0.410$ &	-$0.045$ \\ 
         \gls{allmissing} & Feature & $\secondbest{0.661}$ &	$\secondbest{0.624}$ &	$\best{0.558}$ &	$\secondbest{0.617}$ &	$\best{0.421}$ & $0.548$ &	$0.563$ &	$\secondbest{0.410}$  & $\secondbest{0.342}$ &	$\best{0.155}$ \\
         \bottomrule
    \end{tabularx}
\end{table*}
We analyze the effect of applying two \gls{maug} techniques at different levels of the \gls{mvl} models, similar to Table~\ref{tab_abl:maug} that shows this analysis for average fusion. 
The Table~\ref{tab_supp:maug_gated}, Tables~\ref{tab_supp:maug_cross}, and Tables~\ref{tab_supp:maug_mem} include the results when using gated fusion,  cross-attention fusion, and memory fusion respectively.
When views are missing at input-level, they are imputed, and when views are missing at feature-level, they are ignored.
We notice the same behavior observed for the average, i.e. i) tendency to increase the model robustness to missing data when the \gls{allmissing} technique is used, compared to \gls{sensd} and without \gls{maug} techniques, ii) generalization behavior (increase in performance) of \gls{maug} techniques when there is no missing views.
Nevertheless, we notice that in some cases the \gls{maug} technique impairs the model performance due to the difficulty in estimating the target with missing views.

\section{Results with additional metrics} \label{sec_app:add_metrics}

We assess the predictive performance with alternative metrics: area under the curve (AUC) of the precision-recall plot in classification, and mean average percentage error (MAPE) in regression tasks.
In addition, for assessing the robustness, we use the \gls{prs} presented in \cite{heinrich2023targeted}. 
The \gls{prs} is based on the predictive error with missing views relative to the predictive error in the full-view scenario: 
\begin{equation} 
\text{PRS}(y, \hat{y}_{miss}, \hat{y}_{full}) = \exp{ \left( 1- \cfrac{\text{RMSE}(y, \hat{y}_{miss})}{ \text{RMSE}(y, \hat{y}_{full})} \right) },
\end{equation}
then it is normalized as $PRS =  \min (1, PRS)$.
The results for the \gls{cropharvestB}, \gls{cropharvestM}, \gls{lfmc}, and \gls{pm25} data are in Tables \ref{tab_sup:missing:auc:cropb}, \ref{tab_sup:missing:auc:cropm}, \ref{tab_sup:missing:mape:lfmc}, and \ref{tab_sup:missing:mape:pm25} respectively.
We notice that the model robustness cannot be assessed only with relative robustness metrics, such as \gls{prs}. This is because the relative metrics hide the overall predictive performance. For instance, a horizontal line behavior in Figure~\ref{fig:perc_missing}, such as from I\gls{sensd}-co, will get a \gls{prs} of one, independently of the position of this line on the y-axis (performance). 
Even, in some cases, the prediction shift due to missing views can go towards correcting the original prediction, as shown in F\gls{allmissing}l-co in Figure~\ref{fig:perc_missing}.
In our work, we include these metrics for further analysis, but metrics that can mix these concepts could allow a more succinct analysis.

\begin{table*}[!t]
\caption{Additional results for different cases of missing views (moderate and extreme) in the \textbf{\gls{cropharvestB}} data. We highlight the \best{best} and \secondbest{second best} value in each scenario. The value in parentheses is the number of available views. }\label{tab_sup:missing:auc:cropb}
\centering
\scriptsize
\begin{tabularx}{\textwidth}{l|cCCCC||CCCC} \cmidrule{2-10}
\multicolumn{1}{c}{} & \multicolumn{5}{c||}{AUC value ($\uparrow$)} & \multicolumn{4}{c}{\gls{prs} value ($\uparrow$)} \\ 
\toprule
& (4/4) No & \multicolumn{2}{|c}{(3/4) Only missing} & \multicolumn{2}{|c||}{(1/4) Only available} & 
\multicolumn{2}{c}{(3/4) Only missing} & \multicolumn{2}{|c}{(1/4) Only available} \\
Method & Missing & \multicolumn{1}{|c}{Radar}  & Optical & \multicolumn{1}{|c}{Optical} & Radar &
\multicolumn{1}{c}{Radar}  & Optical & \multicolumn{1}{|c}{Optical} & Radar
\\ \midrule
I\gls{tempd}-co           & $0.920$ &	$0.908$ &	$0.813$ &	$0.772$ &	$0.666$ & $0.958$ &	$0.810$ &	$0.761$ &	$0.606$ \\
I\gls{sensd}-co           & $0.903$ &	$0.899$ &	$0.864$ &	$0.869$ &	$0.736$ & $0.984$ &	$0.930$ &	$\secondbest{0.982}$ &	$\secondbest{0.739}$ \\
F\gls{sensd}-cr           & $0.876$ &	$0.847$ &	$0.798$ &	$0.687$ &	$0.637$ & $0.950$ &	$0.861$ &	$0.783$ &	$0.665$  \\
F\gls{allmissing}l-co           & ${0.930}$ &	$\secondbest{0.925}$ &	$0.903$ &	$0.889$ &	$0.791$ & $\best{0.996}$ &	$\best{0.948}$ &	${0.906}$ &	${0.733}$   \\
FEmbr-sa & $0.923$ &	$0.917$ &	$0.889$ &	$0.866$ &	$0.763$ & $0.987$ &	$0.915$ &	$0.800$ &	$0.651$  \\
ESensI-av           & $0.907$ &	$0.907$ &	$0.879$ &	${0.900}$ &	$\best{0.805}$ & 	$\best{0.996}$ &	$\secondbest{0.943}$ &	$\best{0.995}$ &	$\best{0.863}$  \\ 
\midrule
F\gls{allmissing}-av     &  $\secondbest{0.933}$ &	$\best{0.931}$ &	$\secondbest{0.912}$ &	$\best{0.908}$ &	$0.800$ & $\secondbest{0.991}$ &	$0.890$ &	$0.859$ &	$0.478$ \\
F\gls{allmissing}-ga   & $\best{0.934}$ &	$\best{0.931}$ &	$\best{0.915}$ &	$\secondbest{0.904}$ &	$\secondbest{0.801}$  & $0.986$ &	$0.908$ &	$0.897$ &	$0.645$\\
F\gls{allmissing}-cr    & $0.925$ &	$0.922$ &	$0.903$ &	$0.901$ &	$0.800$ &  $0.991$ &	$0.922$ &	${0.919}$ &	$0.672$  \\ 
F\gls{allmissing}-me    & $0.932$ &	$0.921$ &	$0.908$ &	$0.903$ &	$0.800$ & $0.944$ &	$0.899$ &	$0.904$ &	$0.682$  \\
\bottomrule
\end{tabularx} 
\end{table*}

\begin{table*}[!t]
\caption{Additional results for different cases of missing views (moderate and extreme) in the \textbf{\gls{cropharvestM}} data. We highlight the \best{best} and \secondbest{second best} value in each scenario. The value in parentheses is the number of available views. }\label{tab_sup:missing:auc:cropm}
\centering
\scriptsize
\begin{tabularx}{\textwidth}{l|cCCCC||CCCC} \cmidrule{2-10}
\multicolumn{1}{c}{} & \multicolumn{5}{c||}{AUC value ($\uparrow$)} & \multicolumn{4}{c}{\gls{prs} value ($\uparrow$)} \\ 
\toprule
& (4/4) No & \multicolumn{2}{|c}{(3/4) Only missing} & \multicolumn{2}{|c||}{(1/4) Only available} & 
\multicolumn{2}{c}{(3/4) Only missing} & \multicolumn{2}{|c}{(1/4) Only available} \\
Method & Missing & \multicolumn{1}{|c}{Radar}  & Optical & \multicolumn{1}{|c}{Optical} & Radar &
\multicolumn{1}{c}{Radar}  & Optical & \multicolumn{1}{|c}{Optical} & Radar
\\ \midrule
I\gls{tempd}-co           & $0.962$ &	$0.946$ &	$0.837$ &	$0.850$ &	$0.661$ & $0.893$ &	$0.674$ &	$0.716$ &	$0.649$ \\
I\gls{sensd}-co          & $0.904$ &	$0.869$ &	$0.732$ &	$0.760$ &	$0.595$ & $0.937$ &	$0.759$ &	$0.821$ &	$0.718$  \\
F\gls{sensd}-cr          & $0.928$ &	$0.864$ &	$0.796$ &	$0.779$ &	$0.673$ & $0.905$ &	$0.769$ &	$0.834$ &	$0.695$  \\
F\gls{allmissing}l-co           & 	${0.964}$ &	$0.958$ &	$\secondbest{0.937}$ &	$0.945$ &	$0.856$ &  $\secondbest{0.953}$ &	$\secondbest{0.894}$ &	$\secondbest{0.950}$ &	$\secondbest{0.776}$  \\
FEmbr-sa & 	$0.956$ &	$0.950$ &	$0.911$ &	$0.929$ &	$0.797$ & $0.937$ &	$0.820$ &	$0.761$ &	$0.562$ \\
ESensI-av           &  $0.948$ &	$0.945$ &	$0.912$ &	$\best{0.957}$ & $\best{0.885}$ & $\best{0.971}$ &	$\best{0.895}$ & $\best{0.991}$ & $\best{0.884}$ \\ 
\midrule
F\gls{allmissing}-av     &  $\best{0.967}$ &	$\best{0.962}$ &	$\secondbest{0.937}$ &	$\secondbest{0.954}$ &	$0.864$ & $0.929$ &	$0.810$ &	$0.851$ &	$0.622$  \\
F\gls{allmissing}-ga   & $\best{0.967}$ &	$\best{0.962}$ &	$\best{0.944}$ &	$0.932$ &	$0.854$ & $0.938$ &	$0.838$ &	$0.885$ &	$0.661$ \\
F\gls{allmissing}-cr    & $0.962$ &	$0.958$ &	$0.935$ &	$0.953$ &	$\secondbest{0.875}$ &  $0.949$ &	$0.820$ &	$0.941$ &	$0.703$  \\
F\gls{allmissing}-me    & $\secondbest{0.965}$ &	$\secondbest{0.959}$ &	$0.934$ &	$\secondbest{0.954}$ &	${0.873}$ & $0.942$ &	$0.824$ &	$0.943$ &	$0.738$ \\
\bottomrule
\end{tabularx}
\end{table*}

\begin{table*}[!t]
\caption{Additional results for different cases of missing views (moderate and extreme) in the \textbf{\gls{lfmc}} data. We highlight the \best{best} and \secondbest{second best} value in each scenario. The value in parentheses is the number of available views. }\label{tab_sup:missing:mape:lfmc}
\centering
\scriptsize
\begin{tabularx}{\textwidth}{l|cCCCC||CCCC} \cmidrule{2-10}
\multicolumn{1}{c}{} & \multicolumn{5}{c||}{MAPE value ($\downarrow$)} & \multicolumn{4}{c}{\gls{prs} value ($\uparrow$)} \\ 
\toprule
& (6/6) No & \multicolumn{2}{|c}{(5/6) Only missing} & \multicolumn{2}{|c||}{(1/6) Only available} & 
\multicolumn{2}{c}{(5/6) Only missing} & \multicolumn{2}{|c}{(1/6) Only available} \\
Method & Missing & \multicolumn{1}{|c}{Radar}  & Optical & \multicolumn{1}{|c}{Optical} & Radar &
\multicolumn{1}{c}{Radar}  & Optical & \multicolumn{1}{|c}{Optical} & Radar
\\ \midrule
I\gls{tempd}-co           & $\secondbest{0.157}$ &	$0.293$ &	$0.293$ &	$0.335$ &	$0.335$ & $0.465$ &	$0.465$ &	$0.433$ &	$0.433$   \\
I\gls{sensd}-co           & $0.191$ &	$0.189$ &	$\secondbest{0.230}$ &	$\best{0.249}$ &	$0.292$ & $0.980$ &	$0.819$ &	$0.794$ &	$0.674$ \\
F\gls{sensd}-cr          & 	$0.211$ &	$0.221$ &	$0.287$ &	$0.327$ &	$0.377$ &  $0.966$ &	$0.772$ &	$0.727$ &	$0.627$ \\
F\gls{allmissing}l-co           & $0.406$ &	$0.368$ &	$0.333$ &	$0.378$ &	$0.510$  &  $\best{0.999}$ &	$\best{0.977}$ &	$\best{0.926}$ &	$\best{0.793}$ \\
FEmbr-sa & $0.191$ &	$0.196$ &	$0.265$ &	$0.721$ &	$0.381$ & $0.964$ &	$0.748$ &	$0.073$ &	$0.381$  \\
ESensI-av           &  $0.254$ &	$0.276$ &	$0.278$ &	$\secondbest{0.252}$ & $\best{0.283}$ & ${0.980}$ &	$\secondbest{0.943}$ &	$\secondbest{0.911}$ &	$\secondbest{0.769}$ \\ 
\midrule
F\gls{allmissing}-av     & $0.173$ &	${0.178}$ &	${0.235}$ &	$0.907$ &	$0.623$ & $0.968$ &	$0.791$ &	$0.073$ &	$0.093$  \\
F\gls{allmissing}-ga   & $\best{0.150}$ &	$\best{0.156}$ &	$\best{0.198}$ &	$\secondbest{0.252}$ &	$\secondbest{0.297}$ & $0.971$ &	$0.737$ &	$0.608$ &	$0.513$ \\
F\gls{allmissing}-cr    & $0.217$ &	$0.228$ &	$0.250$ &	$0.405$ &	$0.429$ & $0.920$ &	$0.847$ &	$0.432$ &	$0.383$ \\
F\gls{allmissing}-me    & 	$0.167$ &	$\secondbest{0.165}$ &	$\best{0.198}$ &	$0.268$ &	$0.299$ & $\secondbest{0.981}$ &	$0.806$ &	$0.685$ &	$0.586$  \\
\bottomrule
\end{tabularx}
\end{table*}

\begin{table*}[!t]
\caption{Additional results for different cases of missing views (moderate and extreme) in the \textbf{\gls{pm25}} data. We highlight the \best{best} and \secondbest{second best} value in each scenario. The value in parentheses is the number of available views. }\label{tab_sup:missing:mape:pm25}
\centering
\scriptsize
\begin{tabularx}{\textwidth}{l|cCCCC||CCCC} \cmidrule{2-10}
\multicolumn{1}{c}{} & \multicolumn{5}{c||}{MAPE value ($\downarrow$)} & \multicolumn{4}{c}{\gls{prs} value ($\uparrow$)} \\ 
\toprule
& (3/3) No & \multicolumn{2}{|c}{(2/3) Only missing} & \multicolumn{2}{|c||}{(1/3) Only available} & 
\multicolumn{2}{c}{(2/3) Only missing} & \multicolumn{2}{|c}{(1/3) Only available} \\
Method & Missing & \multicolumn{1}{|c}{Condition}  & Dynamic & \multicolumn{1}{|c}{Dynamic} & Condition. &
\multicolumn{1}{c}{Condition}  & Dynamic & \multicolumn{1}{|c}{Dynamic} & Condition
\\ \midrule
I\gls{tempd}-co           &  $\best{0.439}$ &	$\best{0.439}$ &	$1.242$ &	$1.033$ &	$1.594$ & $0.186$ &	$0.146$ &	$0.185$ &	$0.146$  \\
I\gls{sensd}-co           &  $0.879$ &	$0.878$ &	$1.301$ &	$1.151$ &	$1.644$ & $0.835$ &	$0.690$ &	$0.834$ &	$0.690$  \\
F\gls{sensd}-cr          & $1.003$ &	$1.149$ &	$1.301$ &	$1.147$ &	$1.300$ & $0.783$ &	$0.862$ &	$0.757$ &	$0.832$  \\
F\gls{allmissing}l-co           & $1.281$ &	$0.713$ &	$1.088$ &	$0.747$ &	$1.422$ &    $\best{0.994}$ &	$\best{1.000}$ &	$0.984$ &	$\best{1.000}$    \\
FEmbr-sa &  $0.717$ &	$0.712$ &	$0.976$ &	$1.038$ &	$1.330$ & $0.933$ &	$0.903$ &	$0.591$ &	$0.693$ \\
ESensI-av           & $0.953$ &	$0.754$ &	$1.223$ &	$0.699$ &	$1.561$ & $\secondbest{0.989}$ &	$0.906$ &	$0.991$ &	$0.889$  \\ 
\midrule
F\gls{allmissing}-av     &  $\secondbest{0.581}$ &	$\secondbest{0.561}$ &	$2.891$ &	$4.957$ &	$1.162$ & $0.869$ &	$0.408$ &	$0.080$ &	$0.720$  \\
F\gls{allmissing}-ga   & $0.651$ &	$0.658$ &	$1.030$ &	$\secondbest{0.672}$ &	$\secondbest{1.051}$ & $0.980$ &	$\secondbest{0.995}$ &	$\secondbest{0.996}$ &	$\secondbest{0.992}$ \\
F\gls{allmissing}-cr    & $0.597$ &	$0.614$ &	$\best{0.728}$ &	$\best{0.651}$ &	$\best{0.805}$ & $0.949$ &	$0.680$ &	$0.847$ &	$0.701$ \\
F\gls{allmissing}-me    &  $0.672$ &	$0.653$ &	$\secondbest{0.884}$ &	$0.803$ &	$1.318$ & $0.979$ &	$0.980$ &	$\best{1.000}$ &	$\best{1.000}$  \\
\bottomrule
\end{tabularx}
\end{table*}

\begin{figure*}[!t]
\centering
\subfloat
{\includegraphics[width=0.475\textwidth]{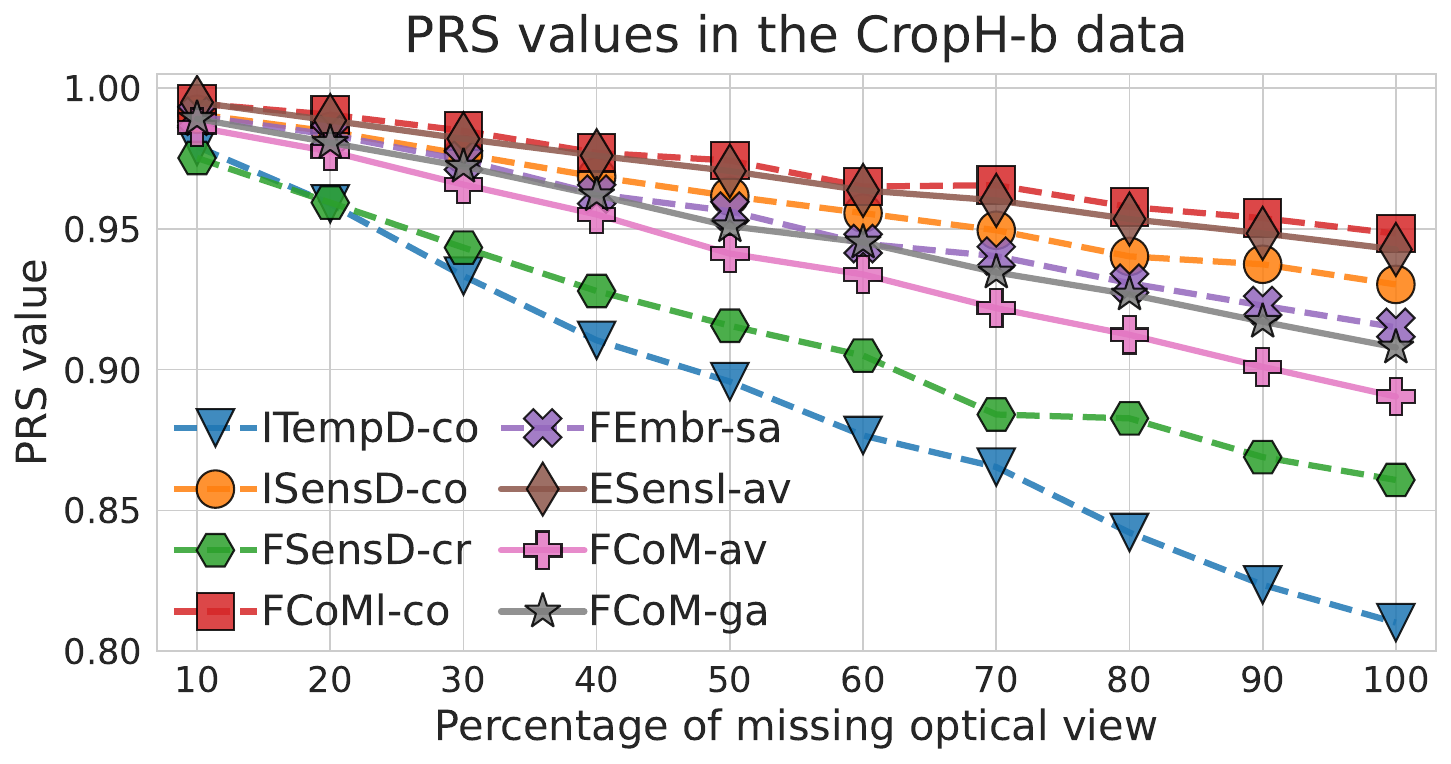} \hfill
\includegraphics[width=0.475\textwidth]{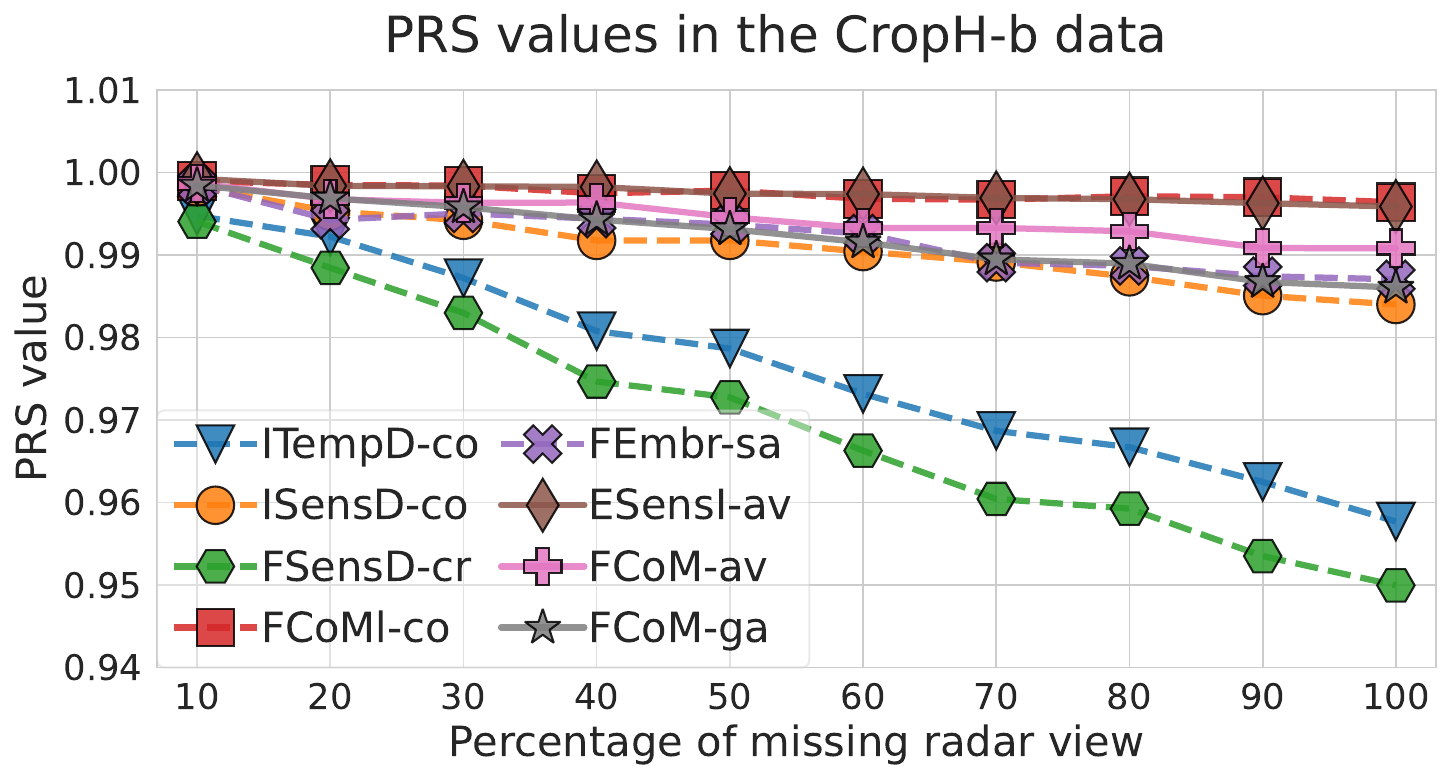}
}
\\
\subfloat
{\includegraphics[width=0.475\textwidth]{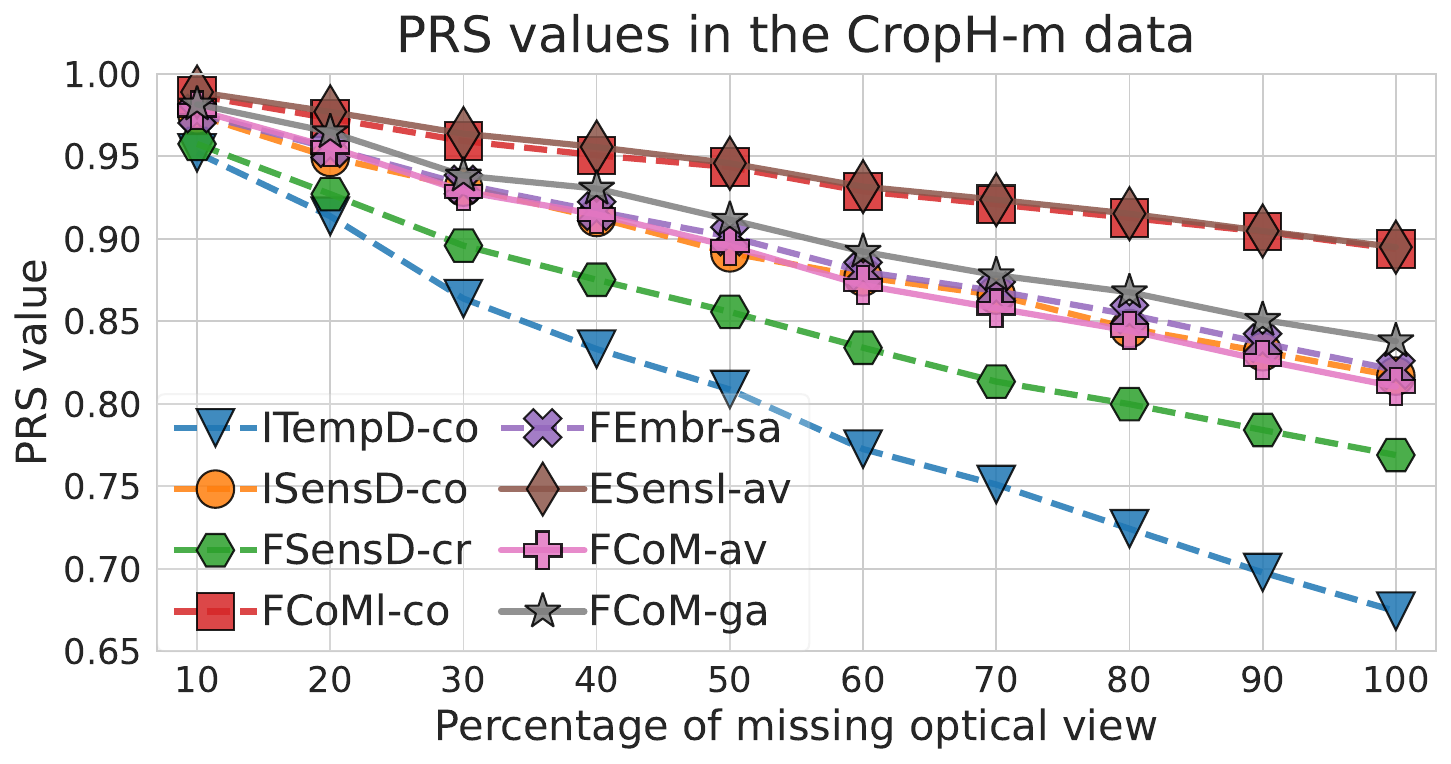} \hfill
\includegraphics[width=0.475\textwidth]{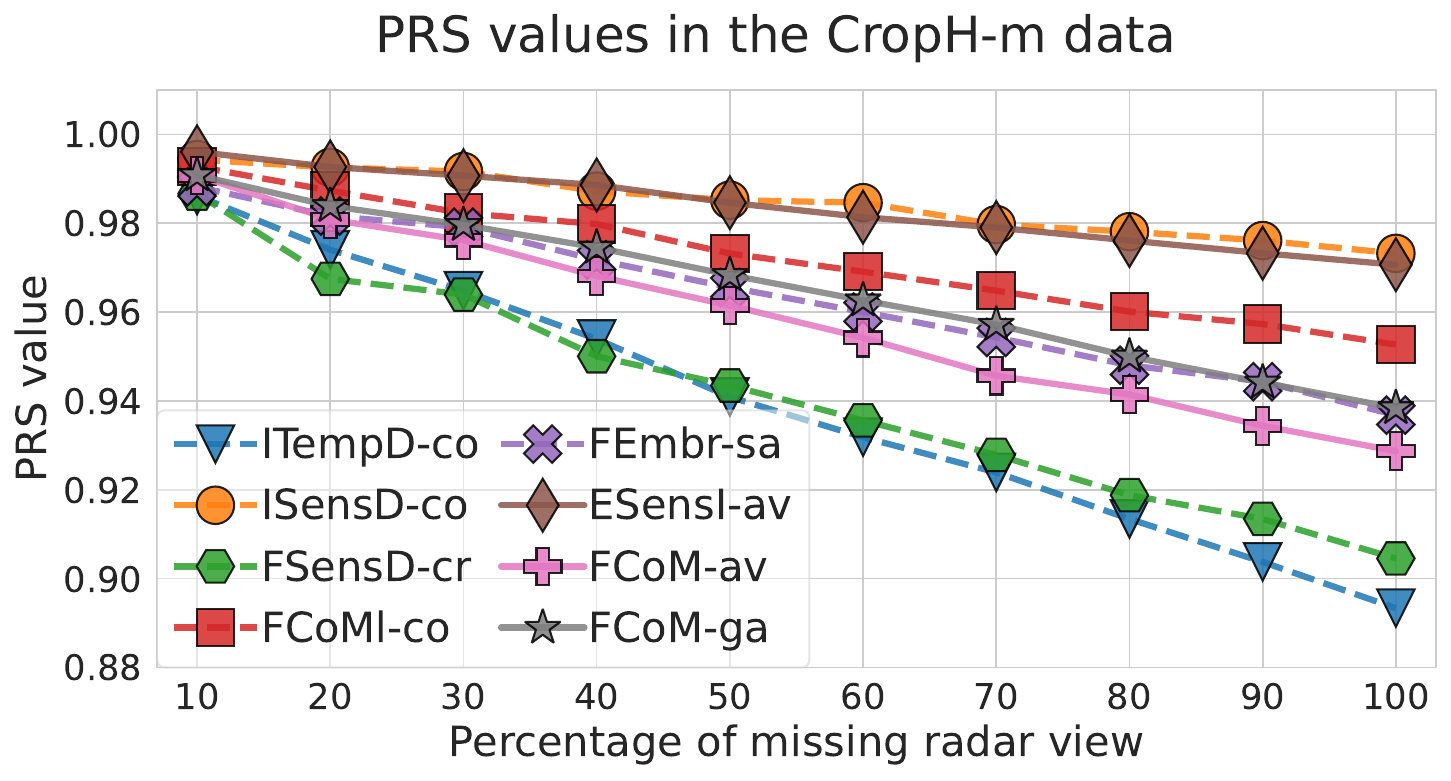}
}
\\
\subfloat
{\includegraphics[width=0.475\textwidth]{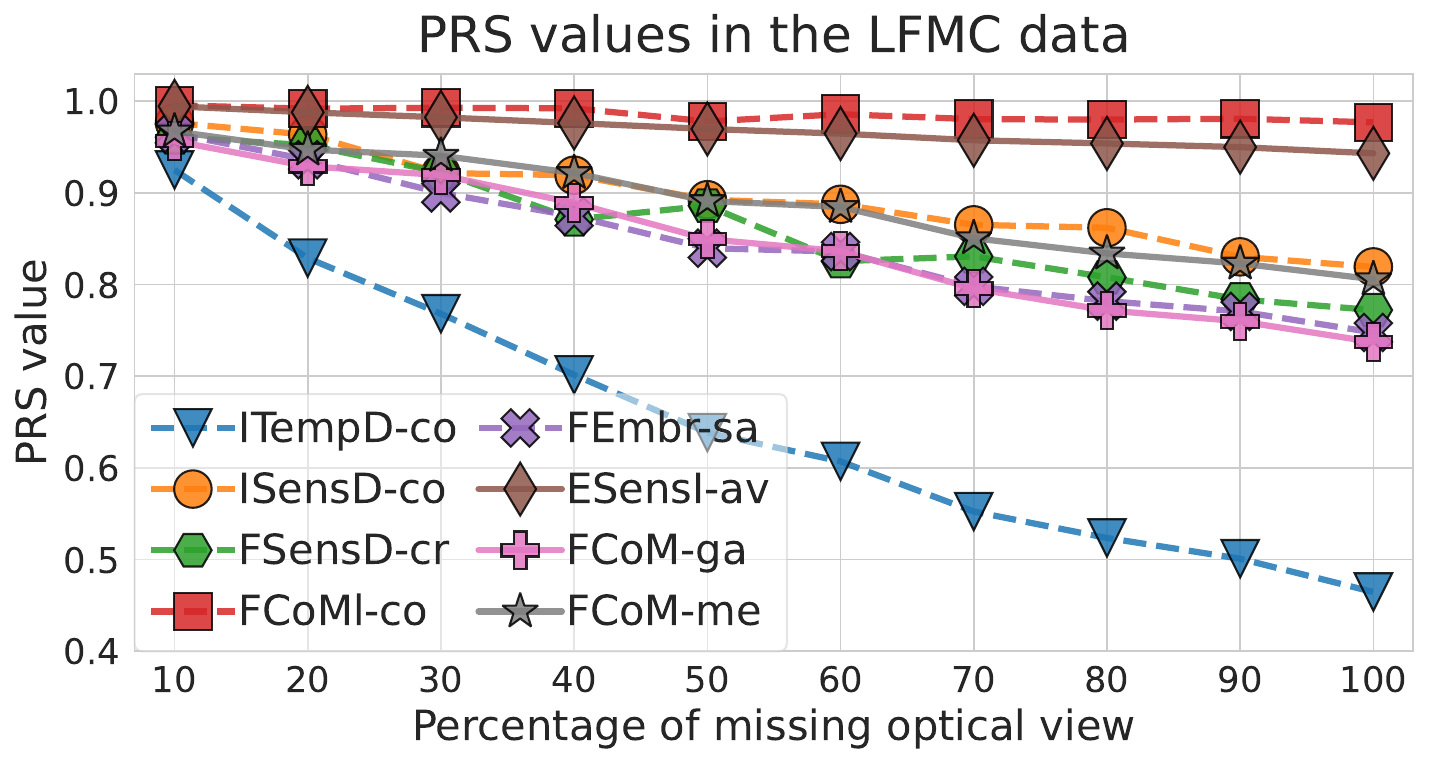} \hfill
\includegraphics[width=0.475\textwidth]{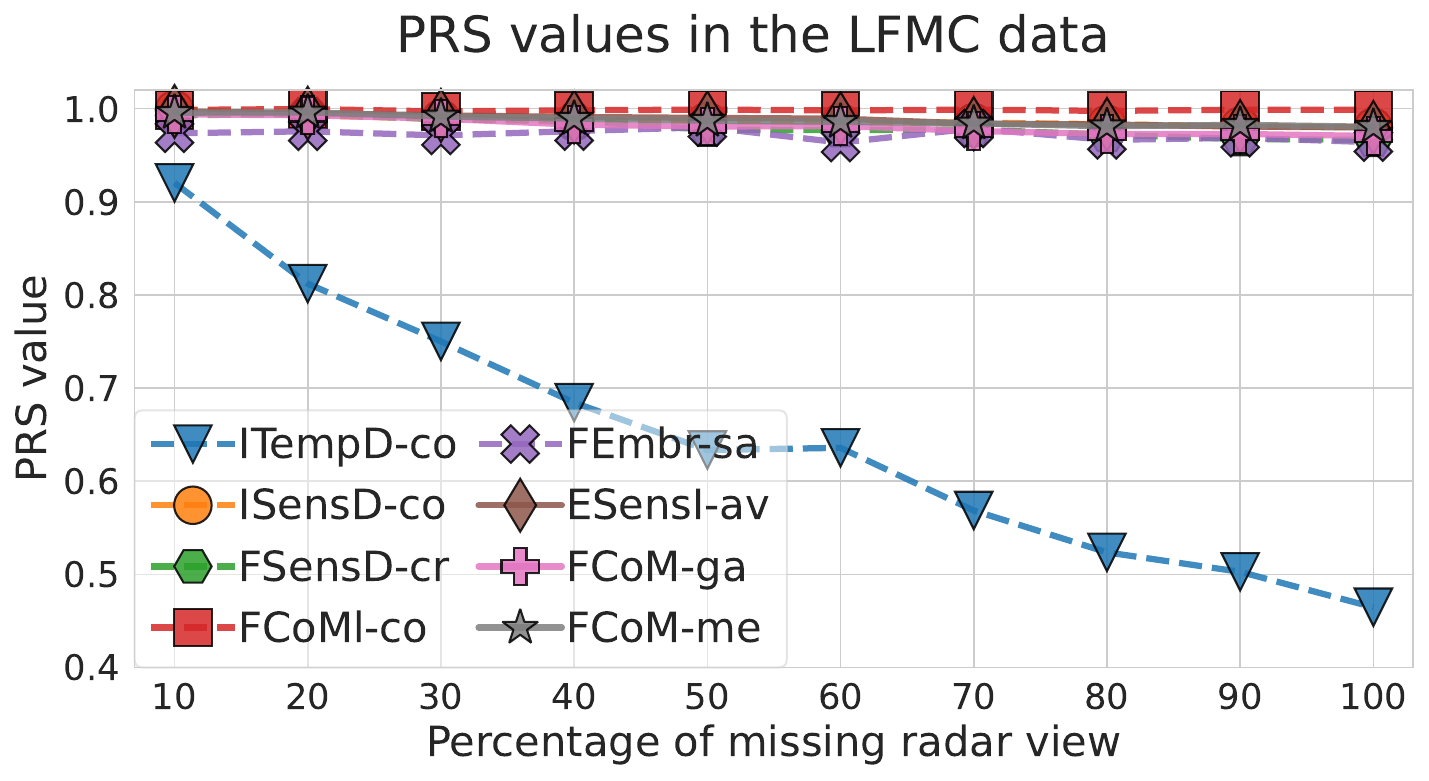}
} 
\\
\subfloat
{\includegraphics[width=0.475\textwidth]{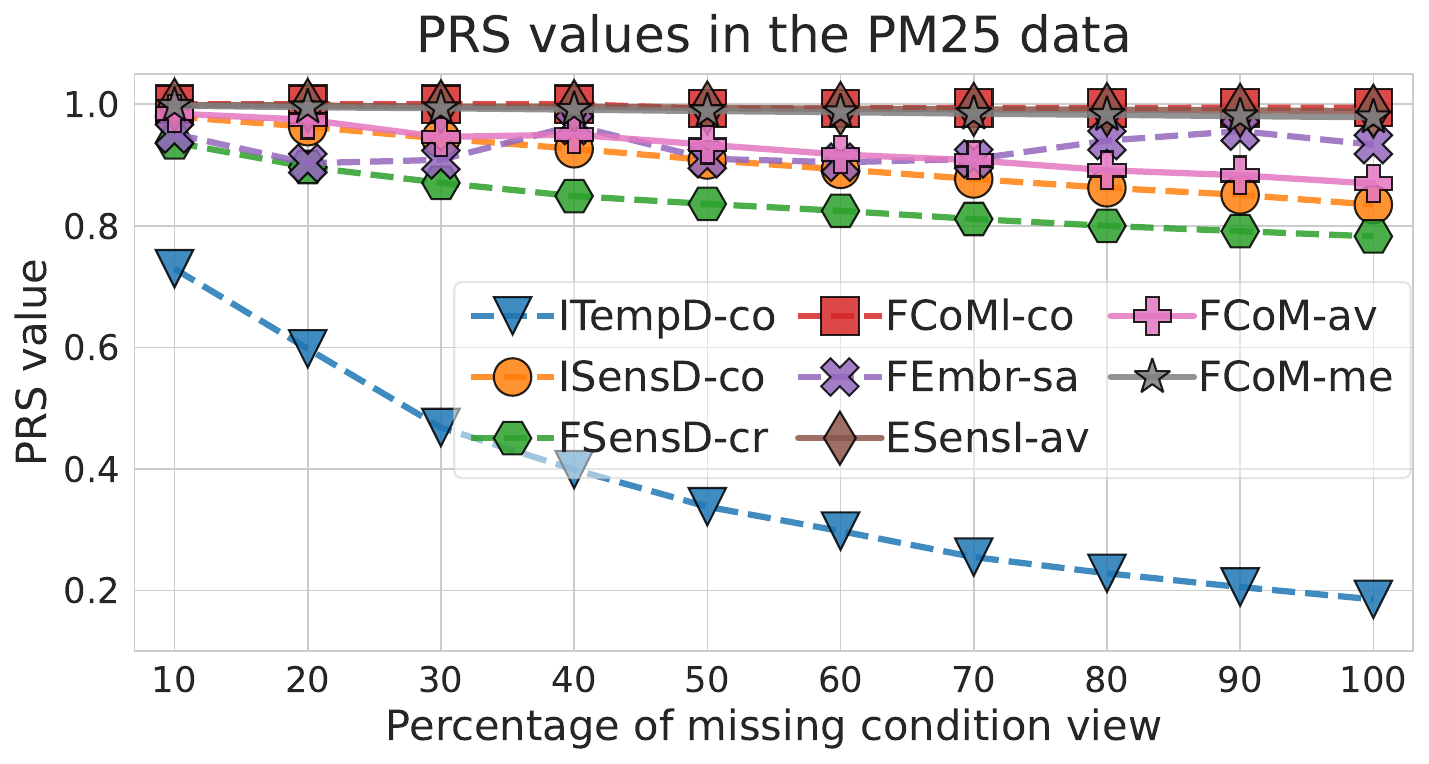} \hfill
\includegraphics[width=0.475\textwidth]{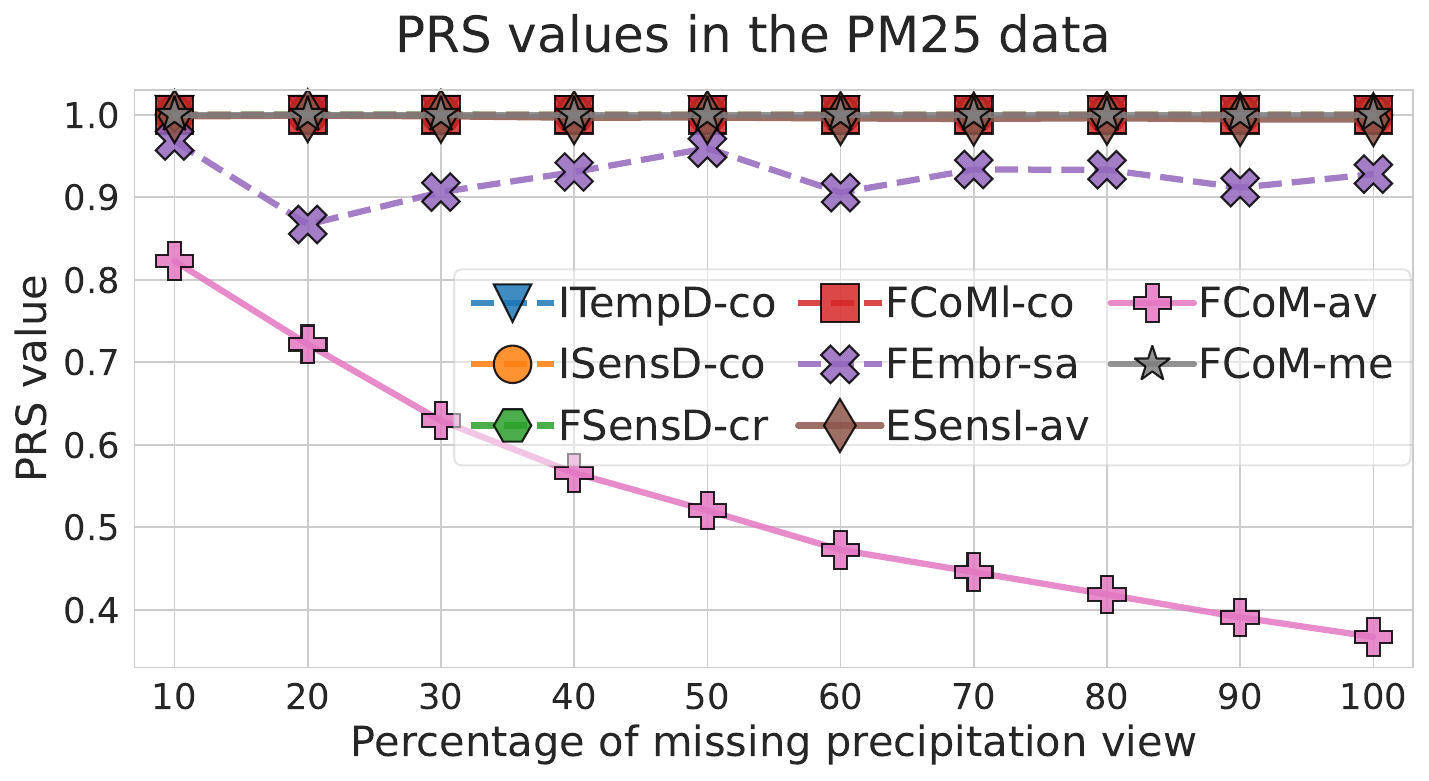}
}
\caption{Predictive robustness when the \textit{top} views are missing at different percentages.}\label{fig_supp:perc_missing:prs}
\end{figure*}

We plot the \gls{prs} value when different number of samples have the top views missing in Figure \ref{fig_supp:perc_missing:prs}.
We notice a different behavior in this relative score compared to results in Figure~\ref{fig:perc_missing}. 
In the \gls{prs} analysis, the best results are obtained by the ESensI-av method followed by F\gls{allmissing}l-co.
The curve of our methods are between the third and fourth best position in this relative score. 
This reflects that, despite the good behavior of our methods in the predictive performance, there is still a gap in reaching the predictive robustness (based on \gls{prs}) of the competing methods, such as the one of ESensI-av.
